\documentclass[sigconf,natbib=true,anonymous=false]{acmart}

\usepackage{times}
\usepackage{soul}
\usepackage{url}
\usepackage[utf8]{inputenc}
\usepackage{graphicx}
\usepackage{amsmath}
\usepackage{amsthm}
\usepackage{booktabs}
\usepackage{algorithm}
\usepackage{algorithmic}
\usepackage[switch]{lineno}
\usepackage{multirow}
\usepackage{tabu}
\usepackage{boldline}
\usepackage{hhline}
\usepackage{amsmath}
\usepackage{xcolor}
\usepackage{pifont}
\usepackage{subfigure}
\usepackage{makecell}
\usepackage{utfsym}
\usepackage{bbding}
\usepackage[normalem]{ulem}
\useunder{\uline}{\ul}{}
\usepackage[inline]{enumitem}
\usepackage{enumitem}
\usepackage{IEEEtrantools}
\usepackage{stmaryrd}
\usepackage{colortbl}
\AtBeginDocument{%
  }

\setcopyright{acmlicensed}
\copyrightyear{2025}
\acmYear{2025}
\acmDOI{XXXXXXX.XXXXXXX}
\acmConference[XXX'25]{The XXX}{XXX,
  2025}{XXX, XXX}
\acmISBN{978-1-4503-XXXX-X/18/06}

\usepackage[most]{tcolorbox}
\begin{document}

\title{LLM-KT: Aligning Large Language Models with Knowledge Tracing using a Plug-and-Play Instruction}

\author{Ziwei Wang$^{1}$, Jie Zhou$^{1*}$, Qin Chen$^{1}$, Min Zhang$^{1}$, Bo Jiang$^{1}$, Aimin Zhou$^{1}$, Qinchun Bai$^{2}$, Liang He$^{1}$
}
\thanks{*Corresponding author, jzhou@cs.ecnu.edu.cn.}
\affiliation{
  \institution{$^{1}$ School of Computer Science and Technology, East China Normal University, China}
  \country{} 
  \institution{$^{2}$ Shanghai Open University, China}
}


\begin{abstract}
The knowledge tracing (KT) problem is an extremely important topic in personalized education, which aims to predict whether students can correctly answer the next question based on their past question-answer records.
Prior work on this task mainly focused on learning the sequence of behaviors based on the IDs or textual information. However, these studies usually fail to capture students' sufficient behavioral patterns without reasoning with rich world knowledge about questions.
In this paper, we propose a large language models (LLMs)-based framework for KT, named \texttt{\textbf{LLM-KT}}, to integrate the strengths of LLMs and traditional sequence interaction models.
For task-level alignment, we design Plug-and-Play instruction to align LLMs with KT, leveraging LLMs' rich knowledge and powerful reasoning capacity.
For modality-level alignment, we design the plug-in context and sequence to integrate multiple modalities learned by traditional methods.
To capture the long context of history records, we present a plug-in context to flexibly insert the compressed context embedding into LLMs using question-specific and concept-specific tokens.
Furthermore, we introduce a plug-in sequence to enhance LLMs with sequence interaction behavior representation learned by traditional sequence models using a sequence adapter.
Extensive experiments show that \texttt{\textbf{LLM-KT}} obtains state-of-the-art performance on four typical datasets by comparing it with approximately 20 strong baselines. 
\end{abstract}

\vspace{-1mm}

\ccsdesc[500]{Computing methodologies~Natural language processing}
\ccsdesc[500]{Applied computing~E-learning}
\ccsdesc[500]{Information systems~Data mining}

\vspace{-2mm}
\keywords{Knowledge Tracing, Intelligent Education, Large Language Models}


\maketitle

\section{Introduction}
Knowledge tracing \cite{abdelrahman2023knowledge,zanellati2024hybrid,shen2024survey} aims to infer students' performance based on their historical question-answer records for personalized education. 
This technique can help teachers and education systems understand the knowledge status of students, such as their skills and forgetting behavior. By doing so, it provides more accurate teaching plans and resources. Effectively solving the knowledge tracing problem can significantly enhance the efficiency of computer-aided education. 

\begin{figure}[!t]
\begin{center}
\includegraphics[width=0.48\textwidth]{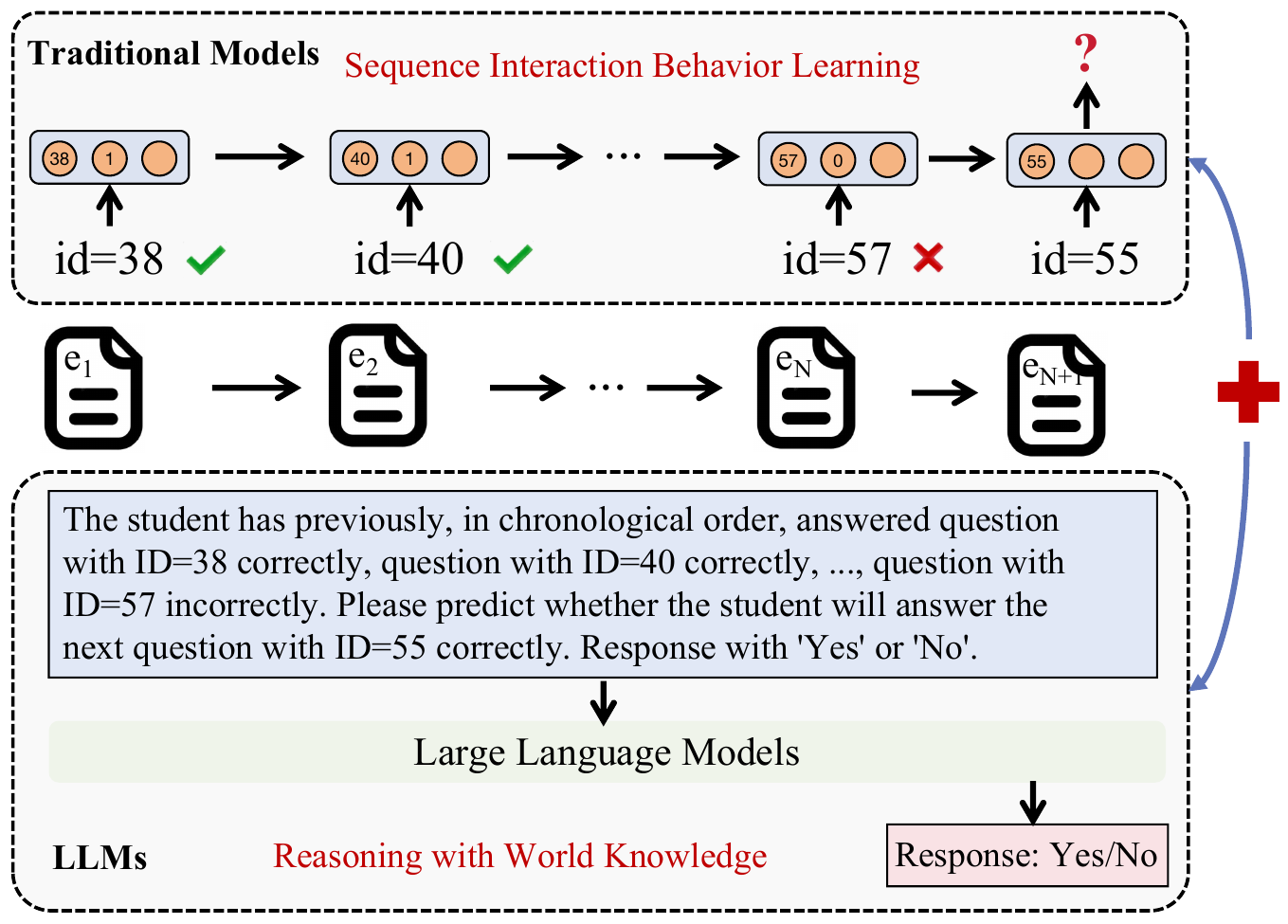}
\end{center}
\caption{The advantages of traditional models and LLMs for knowledge tracing. Traditional models are good at learning the sequence of interaction behavior, while LLMs are good at reasoning with rich world knowledge.} 
\label{fig:intro}
\end{figure}

Traditional deep learning-based models mainly focus on modeling the interaction between questions using IDs (e.g., Question IDs or Concept IDs) to learn the sequence behavior information (See Figure \ref{fig:intro}). 
Sequence learning models (e.g., LSTM \cite{hochreiter1997long} and Transformer \cite{vaswani2017attention}) are utilized to capture the representation of problem-solving records \cite{DKT,SAKT}. 
\citet{DKT+Forget,HawkesKT} integrate the time factor into the model to further learn the sequence information.
Additionally, to learn the relationships among questions and concepts, graph neural networks are adopted \cite{Bi-CLKT,PEBG}. 
These models enhance the interaction representations of an ID sequence using extra knowledge, such as time features and graph structure. 
However, the questions' textual information that contains rich semantic knowledge is not well explored. 

Recently, several studies have incorporated pre-trained language models (PLMs, such as BERT \cite{bert}) into knowledge tracing to model the textual information of the question \cite{BiDKT,tong2020exercise}.
For instance, BiDKT \cite{BiDKT} adapts BERT to trace knowledge by predicting the correctness of randomly masked responses within sequences.  
MLFBK \cite{MLFBK} and LBKT \cite{lstm_bert} leverage BERT to mine complex data relations.  
These methods use the question representations obtained from PLMs to enhance the traditional sequence models.  
Though PLMs are good at natural language understanding, they cannot effectively mine the logic and reasons behind the sequence of questions due to their limited reasoning and world knowledge acquisition abilities.  
Recently, large language models (LLMs) like LLaMA \cite{llama2} have achieved great success in various natural language processing tasks due to their abilities in generation, instruction following, and reasoning.  
In this paper, we aim to integrate LLMs into knowledge tracing to better utilize the world knowledge and powerful reasoning ability of large language models.  

However, there are two primary challenges to applying LLMs for knowledge tracing. 
First (\textbf{C1}), LLMs struggle to capture sequential interaction behaviors from a series of IDs, which reflect students' knowledge states. Large language models interpret question IDs merely as numbers and fail to comprehend user behavior. This often results in splitting IDs into multiple tokens, causing a loss of semantic information associated with those IDs. Experiments indicate that fine-tuning LLaMA can enhance performance, but the gains are limited (refer to Table \ref{table:main results.}). 
Second (\textbf{C2}), LLMs have difficulty accurately capturing the long textual context of comprehensive problem-solving records. The textual content of questions and concepts is crucial for understanding user behavior. However, historical records may include over 200 questions, with each question averaging around 77 tokens. Experimental findings also suggest that the existing strong LLMs like LLaMA and GPT-4o cannot effectively learn students' states from this textual context.

To leverage the strengths of traditional sequence models and LLMs, we propose an LLM-based framework for knowledge tracing, referred to as \texttt{\textbf{LLM-KT}}. Specifically, we create a plug-and-play instruction that incorporates various modalities (such as textual data and IDs) utilizing specific tokens to align LLMs with knowledge tracing from the task level. For \textbf{C1}, we introduce a Plug-in Sequence that translates the embedding of sequence behavior learned from traditional knowledge tracing models into the LLM space. The traditional knowledge tracing model enhances the LLM's capability to comprehend the semantic and interaction information contained in the sequence of IDs. For \textbf{C2}, we propose a Plug-in Context module to substitute the specific token with representations of an extended context for questions. By aligning the compressed context embedding through a context adapter, it successfully captures the semantic textual information of questions. We perform extensive experiments across four standard datasets, with results demonstrating that our model surpasses all strong baselines in the majority of cases. Ablation studies and further analyses also confirm the effectiveness of \texttt{\textbf{LLM-KT}} and its key components. 

The main contribution of this paper can be summarized as follows:
\begin{itemize}[leftmargin=*, align=left]
    \item For task-level alignment, we propose \texttt{\textbf{LLM-KT}} to align LLMs with knowledge tracing using a Plug-and-Play Instruction. We introduce question- and concept-specific tokens to insert embeddings of texts and IDs flexibly.
    \item For modality-level alignment, we design a plug-in sequence to integrate the sequence interaction representations learned by the traditional sequence modeling algorithms with LLMs. Additionally, we present a plug-in context to capture the long textual context of questions.
    \item A series of experiments indicate that our model obtains the new SOTA performance over four benchmark datasets by comparing it with several strong baselines, which indicate the great advantages of our \texttt{\textbf{LLM-KT}} model. 
\end{itemize}

The rest of the paper is presented as follows. First, we review the most related studies in Section \ref{sect:related work}. Then, we introduce the details of our \texttt{\textbf{LLM-KT}} model in Section \ref{sect:methods}. Furthermore, we present the experimental settings of datasets, evaluation metrics, baselines and implementation details in Section \ref{sect:Experimental Settings}. Finally, we give the experimental analysis and conclusions in Section \ref{sect:Experimental Analysis} and \ref{sec:Conclusions and Further Work}.   


\section{Related Work}
\label{sect:related work}
\subsection{Deep learning-based Knowledge Tracing}
Traditional knowledge tracking algorithms are mainly based on machine learning algorithms, such as Bayesian Knowledge Tracing (BKT) \cite{BKT} and Item Response Theory (IRT) \cite{IRT}. 
With the continuous development and progress of neural networks, deep learning-based knowledge tracing algorithms have emerged to model the sequence interaction \cite{DKT,MRT-KT,LPKT,AdaptKT}. 
DKT \cite{DKT}, or Deep Knowledge Tracing, is the first model to apply deep learning to the field of knowledge tracing, which learns the features of students' historical problem-solving records using Long Short-Term Memory (LSTM). 
SAKT \cite{SAKT} utilized the self-attention mechanism to address the problem of insufficient generalization ability existing in the processing of sparse data. 
AKT \cite{AKT} further introduced a new monotonic attention mechanism and the classic Rasch-model in psychometrics to better understand students' knowledge mastery status and learning processes.
BEKT \cite{BEKT} proposed a multi-layer bidirectional transformer encoder with a self-attention mechanism and bidirectional analysis, to understand the student's past learning logs.
\citet{DKVMN} proposed a new structure called Dynamic Key-Value Memory Networks (DKVMN), which can utilize the relationships between underlying concepts and directly output the mastery level of each concept by students.  

To further evaluate the time aspect, DKT-Forget \cite{DKT+Forget} enhances DKT by translating the time interval into a numerical value. This value, along with learning interaction data like answering questions, is fed into the neural network. In contrast, HawkesKT \cite{HawkesKT} leverages the intensity function and mechanisms of the Hawkes process to measure the triggering effects of events across different time points. This approach clarifies how learning events temporally influence the probability of subsequent occurrences and the knowledge state.
Addressing limitations in the learning process, which is vital for KT tasks, LPKT \cite{LPKT} assesses students’ knowledge states by modeling their learning journey, capturing knowledge gains while also considering the phenomenon of forgetting. Simultaneously, \citet{ProKT} offers a novel perspective in the KT field by developing the Progressive Knowledge Tracing model. This model emphasizes the learning journey through students’ sequential thought processes and divides it into three relatively independent, yet progressively advanced stages: concept mastery, question-solving, and answering behavior, effectively modeling the transition from abstract reasoning to concrete responses.

Furthermore, graph neural networks are used to model the relationships between different questions or knowledge points in the field of knowledge tracing \cite{GKT,Bi-CLKT,PEBG,DGEKT}. 
GKT \cite{GKT} constructs a knowledge graph based on knowledge points or questions, and utilizes Graph Neural Networks (GNNs) to explore and take advantage of these underlying relational structures. 
BI-CLKT \cite{Bi-CLKT} designs a two-layer comparative learning scheme on an ``exercise-to-exercise" (E2E) relational subgraph for node-level and graph-level contrastive learning to get discriminative representations of exercises and concepts. Additionally, two variants with different prediction layers (RNN and memory-augmented neural networks) are explored to improve representations.
PEBG \cite{PEBG} puts forward a pre-training embedding method through a bipartite graph (PEBG), leveraging edge information (including question difficulty, explicit question-skill relationships, implicit question similarity, and skill similarity) to learn low-dimensional embeddings for each question. 
DGEKT \cite{DGEKT} innovatively constructs a dual graph structure of students' learning interactions, using a concept association hypergraph and a directed transition graph to capture heterogeneous relationships. Additionally, it employs online knowledge distillation to adaptively combine the dual graph models, forming a stronger ensemble teacher model for enhanced modeling ability.

\begin{figure*}[!t]
\begin{center}
\includegraphics[width=1\textwidth]{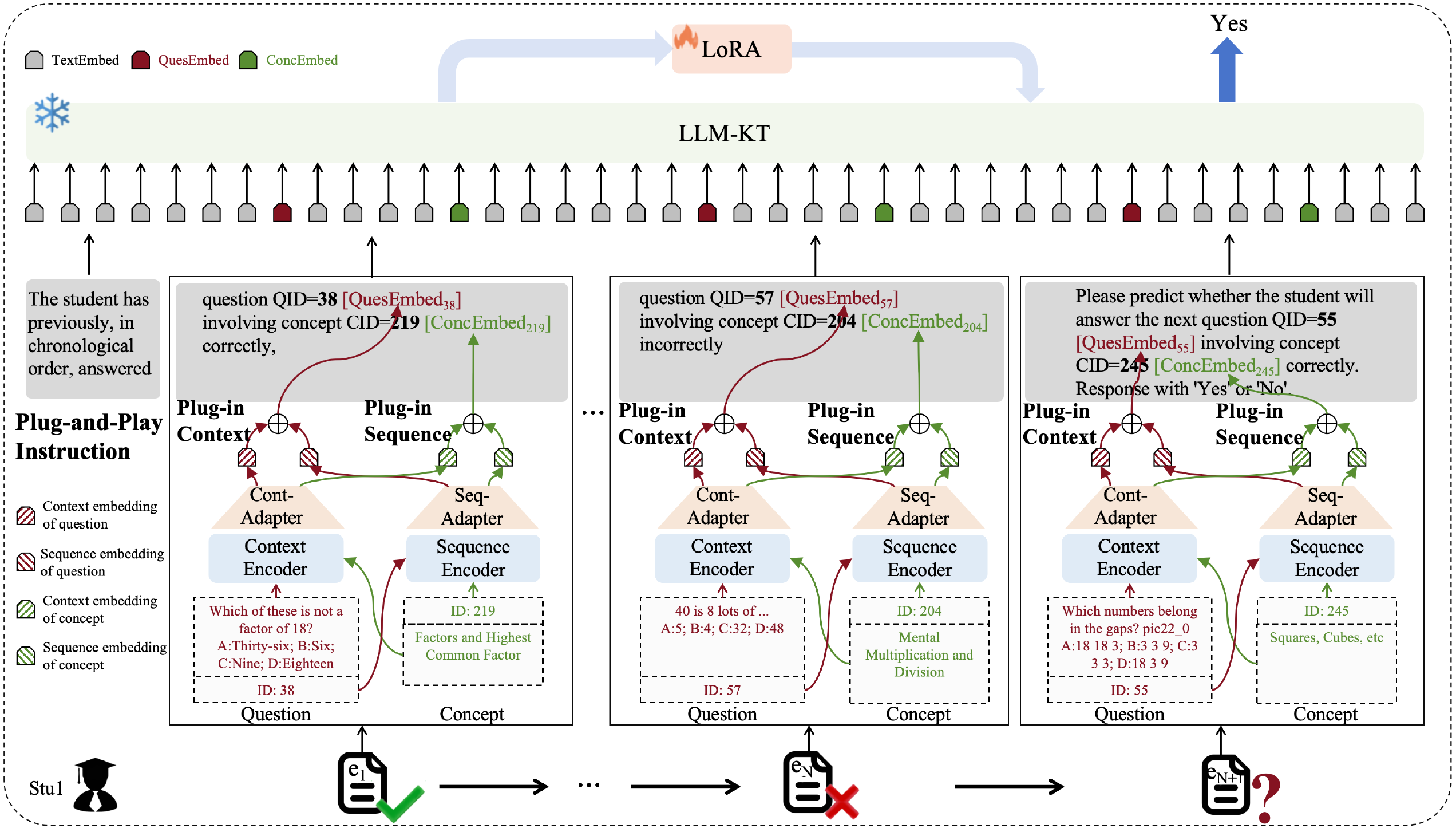}
\end{center}
\caption{The framework of \texttt{\textbf{LLM-KT}}. We propose a Plug-and-Play Instruction to combine the strengths of LLMs and traditional sequence models for knowledge tracing by inserting multiple modalities into LLMs. Particularly, we design a Plug-in Context module to capture the long context of students' problem-solving records. Then, we introduce the Plug-in Sequence to align the sequence interaction representation learned by the traditional model with LLMs.} 
\label{fig:main}
\end{figure*}

\subsection{PLMs-Enhanced Knowledge Tracing}
In the field of knowledge tracing, Pre-trained Language Models (PLMs), such as BERT \cite{bert}, RoBERT \cite{RoBERT}, are used to enhance the semantic representation for knowledge tracing \cite{BiDKT,MLFBK,tong2020exercise,DCL4KT-A}. 
For instance, BiDKT \cite{BiDKT} adapts BERT to trace knowledge by predicting the correctness of randomly masked responses within sequences.
MLFBK \cite{MLFBK} leverages the power of BERT to mine latent relations among multiple explicit features, such as individual skill mastery, students' ability profiles, and problem difficulty.
Furthermore, \citet{tong2020exercise} proposes a hierarchical exercise feature enhanced knowledge tracing framework that utilizes BERT to generate exercise text embeddings and then feeds them into three systems (KDES, DFES, and SFES) to extract knowledge distribution, semantic features, and difficulty features. These hierarchical features are concatenated with student responses and input into a sequence model, aiming to improve knowledge tracing by comprehensively considering the diverse attributes of exercises.
Moreover, LBKT \cite{lstm_bert} combines the strengths of BERT for capturing complex data relations and LSTM for handling long sequences, enhancing performance on data with over 400 interactions. 
However, the integration of LLMs with knowledge tracing has not been explored well.

\subsection{Context-aware Knowledge Tracing}
The context information (such as textual features in the questions and concepts) contains a wealth of semantic knowledge, which can help reduce the cold-start phenomenon of knowledge tracing. 
Several studies utilized the context information to enhance traditional deep learning models \cite{RKT,EERNN}. 
For example, RKT \cite{RKT} used the textual information of the questions to capture relations between exercises. 
EERNN \cite{EERNN} and EKT \cite{EKT} considered the text of the questions to learn a good question representation for knowledge tracing. 
Additionally, \citet{PEBG} proposed a pre-training method called PEBG, which learns question embeddings with rich relational information using the bipartite graph of question-skill relations. 
Moreover, \citet{DCL4KT-A} proposed a difficulty-centered contrastive learning method based on the question representations using BERT. 

Unlike previous research, we propose \texttt{\textbf{LLM-KT}} to combine the great advantages of LLMs and traditional sequence learning models for knowledge tracing. 
We design a plug-and-play instruction to align the context and sequence representations with LLMs.




\section{Our Proposed Method}
\label{sect:methods}
In this section, we propose our \texttt{\textbf{LLM-KT}} framework, a novel LLM-based framework specifically designed to handle knowledge tracing tasks (See Figure \ref{fig:main}).  
To align the LLMs with knowledge tracing from the task level, we design a plug-and-play instruction with specific tokens to flexibly integrate multiple modalities (e.g., long context and IDs) into LLMs. 
In this way, we combine the advantages of LLMs, which contain rich knowledge and strong reasoning capacities, with traditional sequence models that excel at learning sequence interaction behavior.  
For the modality level, we first propose a plug-in context to capture the long context of the sequence of questions using a context encoder and align the vector representations with LLMs via a context adapter.  
Then, we introduce a plug-in sequence to enhance the LLMs with the sequence interactive representation learned from traditional sequence models for knowledge tracing.

Knowledge tracing involves predicting whether a student will answer a new question correctly based on their historical question-answer records. Formally, given a student's exercise history as $H=(e_1, e_2, ..., e_i, ..., e_{N})$, where $N$ is the number of historical exercises. Here, $e_i = (q_i, a_i)$, where $q_i$ represents the information of the $i$-th question the student answered and $a_i$ indicates the student's response to this question ($a_i = 1$ means the student answered correctly, and $a_i = 0$ means the student answered incorrectly). The goal of the task is to predict the value of $a_{N+1}$ (also defined as $y$) when the student answers $q_{N+1}$.


\subsection{Plug-and-Play Instruction}
To utilize the rich world knowledge and reasoning capacity, we design a Plug-and-Play Instruction for \texttt{\textbf{LLM-KT}} to guide LLMs in helping the model accurately capture changes in student learning behaviors and knowledge states.  
Specifically, we develop instructions with question-specific and concept-specific slots to seamlessly integrate knowledge from various modalities, including IDs and contextual information. 
We insert specific tokens and replace the token embeddings with representations of long context and IDs that contain rich semantic and interaction information.
The Plug-and-Play Instruction organizes data into (input, output) pairings using designated tokens. 
The input consists of a student's prior question-answer records along with information about the current question requiring an answer. The output indicates the LLM's predicted evaluation of the student's expected answer's accuracy. Specifically, the instruction with input prompt $x$ and output $y$ is structured as follows:
\\
\noindent\par
\noindent\resizebox{.99\columnwidth}{!}{
\begin{tcolorbox}[left=1mm,right=1mm,top=1mm,bottom=1mm]
\textbf{Input $x$.} \\
The student has previously, in chronological order, answered \textcolor{blue}{\textbf{\{}}question QID=38 [\textcolor{red}{QuesEmbed$_{38}$}] involving concept CID=219 [\textcolor[RGB]{0,205,102}{ConcEmbed$_{219}$}] correctly, ..., question QID=57 [\textcolor{red}{QuesEmbed$_{57}$}] involving concept CID=204 [\textcolor[RGB]{0,205,102}{ConcEmbed$_{204}$}] incorrectly\textcolor{blue}{\textbf{\}}$_\text{HistoryRecord}$}. \\
Please predict whether the student will answer the next \textcolor{cyan}{\textbf{\{}}question QID=55 [\textcolor{red}{QuesEmbed$_{55}$}] involving concept CID=245 [\textcolor[RGB]{0,205,102}{ConcEmbed$_{245}$}] correctly\textcolor{cyan}{\textbf{\}$_\text{TargetQues}$}}. Response with 'Yes' or 'No'. Response:

\textbf{Output $y$.} \\
Yes/No
\end{tcolorbox}
}

In this template $x$, \textcolor{blue}{\textbf{\{\}}$_\text{HistoryRecord}$} represents the student's historical question-answer records, which consist of a series of (question, answer) pairs. Each question includes specific attributes such as question ID and knowledge concepts, and the answer indicates whether the student answered correctly or incorrectly; for example, ``question QID=38 [\textcolor{red}{QuesEmbed$_{38}$}] involving concept CID=219 [\textcolor[RGB]{0,205,102}{ConcEmbed$_{219}$}] correctly.'' Here, [\textcolor{red}{QuesEmbed$_{38}$}] and [\textcolor[RGB]{0,205,102}{ConcEmbed$_{219}$}] mean the specific tokens of question $38$ and concept $219$, where the embeddings are learned by Plug-in Context and Plug-in Sequence. \textcolor{cyan}{\textbf{\{\}$_\text{TargetQues}$}} represents the information about the target question to be predicted without indicating whether the answer is correct or incorrect. This approach allows the LLMs to understand the context of each question and the student's performance related to it.

Then, we input the $x$ into an LLM $\mathcal{M}_{\theta}$ to predict the label $\hat{y}$, where $\theta$ is the trainable parameters of the model. $\hat{y}$=``Yes'' if the student answer the target question $q_{N+1}$ correctly, otherwise $\hat{y}$=``No''. 
In the training phase, we use the language model with cross-entropy loss as the training objective.
Specifically, we adopt Low-Rank Adaptation (LoRA) \cite{LoRA}, which is a popular parameter-efficient fine-tuning method. The core idea of LoRA is to adapt model parameters through low-rank decomposition. 
LoRA adds two low-rank matrices, $A$ and $B$, to the model's weight matrix. 
During fine-tuning, only these two low-rank matrices are updated ($\theta=\{A, B\}$), while the original weight matrix remains unchanged. 
This approach significantly reduces the number of parameters that need to be updated, thereby lowering computational and storage costs.

Furthermore, the question representation [\textcolor{red}{QuesEmbed$_{QID}$}] = $e_{Ques} \in \mathbb{R}^{d^e}$ and concept representation [\textcolor[RGB]{0,205,102}{ConcEmbed$_{CID}$}] = $e_{Conc} \in \mathbb{R}^{d^e}$ consist of context and sequence embeddings that are learned by Plug-in Context $f_{\text{cont}}$ (Section \ref{sect:plug-in context}) and Plug-in Sequence $f_{\text{seq}}$ (Section \ref{sect:plug-in sequence}), where ${d^e}$ is the dimension of LLM's embedding layer.
\begin{equation}
\label{equ:combine}
\begin{aligned}
    e_{Ques} = g(f_{\text{cont}}(QText), f_{\text{seq}}(QID)) \\
    e_{Conc} = g(f_{\text{cont}}(CText), f_{\text{seq}}(CID))
\end{aligned}
\end{equation}
where $g(a,b)$ is a function to combine $a$ and $b$, such as concatenation, average and addition. $QText$ and $CText$ are the textual information of the question and concept, $QID$ and $CID$ are the ids of question and concept.

In the inference phase, we calculate the probability distribution as follows:
\begin{equation}
p(\hat{y}=``\mathrm{Yes}"|x)=\frac{e^{\mathcal{M}_{\theta}(``\mathrm{Yes}"|x)}}{e^{\mathcal{M}_{\theta}(``\mathrm{Yes}"|x)}+e^{\mathcal{M}_{\theta}(``\mathrm{No}"|x)}}
\end{equation}
where $\mathcal{M}_{\theta}(``\mathrm{Yes}"|x)$ and $\mathcal{M}_{\theta}(``\mathrm{No}"|x)$ are the output probabilities of token ``Yes'' and ``No'' of $\mathcal{M}_\theta$.

\subsection{Plug-in Context}
\label{sect:plug-in context}
Due to the challenge of processing excessively long texts with large models, it is impractical to model the complete history of all questions. 
Thus, we propose a Plug-in Context $f_{\text{cont}}$ to model the textual information of question and concept. Particularly, we adopt a context encoder to model the text, and then a context adapter is used to align the representation with LLMs.

\subsubsection{Context Encoder}
We leverage a context encoder to encode question and knowledge concept texts into vector representations. 
Here, we use a pre-trained language model (e.g., LLaMA2 \cite{llama2}, BERT \cite{bert}, and all-mpnet-base-v2 \cite{Mpnet}) as the context encoder to obtain the textual question representation $r_{QText} \in \mathbb{R}^{d^t}$ and concept representation $r_{QText} \in \mathbb{R}^{d^t}$, where $d^t$ is the hidden dimension of context encoder. 
These representations help knowledge tracing models capture the semantic relationships between questions and knowledge concepts.
\begin{equation}
\begin{aligned}
    r_{QText} = \mathrm{ContextEncoder}(QText)  \\
    r_{CText} = \mathrm{ContextEncoder}(CText)
\end{aligned}
\end{equation}

\subsubsection{Context Adapter}
Then, we align these representations with the semantic space of the large model to ensure compatibility and effective knowledge tracing.
To achieve this alignment, we design a context adapter to map the text representation learned by the context encoder to LLMs. 
\begin{equation}
\begin{aligned}
    h_{QText} = \mathrm{ContAdapter}(r_{QText}) \\
    h_{CText} = \mathrm{ContAdapter}(r_{CText})
\end{aligned}
\end{equation}
where the $h_{QText} \in \mathbb{R}^{d^e}$ and $h_{CText} \in \mathbb{R}^{d^e}$ are the representations of question and concept after alignment.
Particularly, we used a simple multi-layer perceptron (MLP) layer as a context adapter to translate the space.

\subsection{Plug-in Sequence}
\label{sect:plug-in sequence}
Large language models have great abilities in natural language understanding and generation by training on large-scale text data. 
To integrate new modalities like ID sequence into LLMs, we propose a Plug-in Sequence $f_{\text{seq}}$ to better capture the interactive information.
In this way, we can utilize the strengths of LLMs and traditional models to learn both semantic and interaction behavior simultaneously.
Similarly, we adopt traditional models as a sequence encoder to learn the sequence representation. Then, a sequence adapter is used to convert the sequence representation into the space of LLMs.

\subsubsection{Sequence Encoder}
We treat the sequence of IDs as a new modality to be injected into LLMs. We adapt existing traditional sequence learning models (e.g., DKT \cite{DKT}, AKT \cite{AKT}) for knowledge tracing to learn question ID embeddings $r_{QID} \in \mathbb{R}^{d^s}$ and concept ID embeddings $r_{CID} \in \mathbb{R}^{d^s}$, where $d^s$ means the dimension of sequence encoder. 
These models are good at learning the sequence interaction based on the question IDs and concept IDs.
\begin{equation}
\begin{aligned}
    r_{QID} = \mathrm{SeqEncoder}(QID)  \\
    r_{CID} = \mathrm{SeqEncoder}(CID)
\end{aligned}
\end{equation}

\subsubsection{Sequence Adapter}
However, due to the semantic space of the traditional model being quite different from that of the large model, we introduce a sequence adapter to align the ID representations from the traditional knowledge tracer with LLMs. This approach ensures better integration and enhances the performance of the knowledge tracing task.
\begin{equation}
\begin{aligned}
    h_{QID} = \mathrm{SeqAdapter}(r_{QID})  \\
    h_{CID} = \mathrm{SeqAdapter}(r_{CID})
\end{aligned}
\end{equation}
where $h_{QID} \in \mathbb{R}^{d^e}$ and $h_{CID} \in \mathbb{R}^{d^e}$ are the aligned representations of question ids and concept ids.

\begin{table*}[t]
\centering
\caption{The statistical information of the datasets. QID and CID mean the ID of the question and concept. KCs means the number of knowledge concepts. QuesCont and ConcCont represent the context of the question and concept.}
\label{table:statistic of datasets}
\vspace{-1mm}
\setlength{\tabcolsep}{3.0mm}{
\begin{tabular}{lcccccccc} 
\hlineB{4}
Dataset & Students & Questions & KCs & Interactions & QID & CID & QuesCont & ConcCont  \\ \hline
Assist2009 & 4,151 & 16,891 & 110 & 325,637  & \ding{51} & \ding{51} & \ding{55} & \ding{51} \\ 
Assist2015 & 19,840 & - & 100 & 683,801  & \ding{55} & \ding{51} & \ding{55} & \ding{55}\\ 
Junyi & 1,000 & 834 & - & 972,855 & \ding{51} & \ding{55} & \ding{55} & \ding{55} \\ 
Nips2020 & 5,310 & 110 & 17 & 428,596  &  \ding{51} &\ding{51} &\ding{51} &\ding{51} \\ \hlineB{4}
\end{tabular}
}
\vspace{-1mm}
\end{table*}

\section{Experimental Settings}
\label{sect:Experimental Settings}
In this section, we first introduce the datasets and evaluation in Section \ref{sect:Datasets and Evaluation}. Then, we list the baselines in Section \ref{sect:Baseines} and provide the implementation details in Section \ref{sect:Implementation Details}.

\subsection{Datasets and Evaluation}
\label{sect:Datasets and Evaluation}
\subsubsection{Datasets} 
To evaluate the effectiveness of our \texttt{\textbf{LLM-KT}}, we conduct experiments on four commonly used benchmark datasets for knowledge tracing. The statistical information of these datasets is listed in Table \ref{table:statistic of datasets}.

\begin{itemize}[leftmargin=*, align=left]
    \item ASSISTments2009 (Assist2009) \cite{AKT} collects the exercises of 4151 students during the 2009 to 2010 school year. The same as \citet{AKT}, we use the skill builder data version of this dataset. 
    To ensure the validity of the data, we only retain those records where both the skill\_name and skill\_id fields are not empty. 
    \item ASSISTments2015 (Assist2015) \cite{AKT} comprises responses from students on 100 distinct questions. Different from Assist2009, this dataset does not provide metadata of questions.
    \item Junyi Academy (Junyi) \cite{Edudata} is provided by Junyi Academy - the premier online learning platform in Taiwan, consisting of over 16 million exercise attempt logs. 
    These logs are contributed by more than 72,000 students, spanning a year, specifically from August 2018 to July 2019. 
    We use the dataset provided by \citet{Edudata}, which is processed specifically for knowledge tracing.
    \item NeurIPS 2020 Education Challenge (Nips2020) \cite{Nips2020} is released by the NeurIPS 2020 Education Challenge. 
    In this paper, we use the datasets from Challenge Task 3 \& 4 and extract the records of the top 150 most frequently appearing questions. 
    Note that, to obtain the textual information of the questions, we convert the figures into text manually.
\end{itemize}

\subsubsection{Evaluation} 
Following \cite{MRT-KT,AKT,LPKT}, we use two widely-used metrics: Area Under the Curve (AUC) and Accuracy (ACC) to evaluate the effectiveness of our model. 

\subsection{Baselines}
\label{sect:Baseines}
In our research, we compare our proposed approach with several strong baseline methodologies to assess its efficacy and performance. We split these baselines into four parts: deep-learning (DL)-based, pre-trained language models (PLMs)-based, context-aware and LLMs-based methods.

\subsubsection{DL-based Methods} 
DL-based methods learn the interactions among students' records effectively by taking the relationships and times into account. Here, we select 7 typical baselines as follows: 
\begin{itemize}[leftmargin=*, align=left]
    \item \textbf{DKT} \cite{DKT} uses RNNs(\cite{RNN} to model temporal dependencies in student learning, capturing the evolution of knowledge states.
    \item \textbf{DKVMN} \cite{DKVMN} implements a dynamic key-value memory network, where static matrices store knowledge concepts and dynamic matrices update mastery levels, enhancing the modeling of concept relationships.
    \item \textbf{SAKT} \cite{SAKT} employs a self-attention mechanism(\cite{vaswani2017attention}) to identify key knowledge concepts (KCs) from past interactions.
    \item \textbf{AKT} \cite{AKT} utilizes a monotonic attention mechanism to build context-aware representations of student interactions, capturing performance over appropriate time scales.
    \item \textbf{LPKT} \cite{LPKT} models the learning process by formalizing learning cells and incorporating gates for managing retention and forgetting over time.
    \item \textbf{LBKT$^\dag$} \cite{LBKT} analyzes the interplay of learning behaviors (e.g., speed, attempts, hints) and uses a forgetting factor to update learners’ knowledge states.
    \item \textbf{MRT-KT} \cite{MRT-KT} employs a multi-relational transformer with a novel relation encoding scheme to model fine-grained interactions between question-answer pairs in knowledge tracing.
\end{itemize}

\subsubsection{PLMs-based Methods}
PLMs-based methods improve the performance of knowledge tracing via the rich knowledge and powerful natural language understanding of PLMs. Here, we adopt the following baselines:
\begin{itemize}[leftmargin=*, align=left]
    \item \textbf{LBKT$^\ddag$} \cite{lstm_bert} addresses long-sequence data in knowledge tracing by integrating a BERT-based architecture with Rasch model embeddings for difficulty levels and an LSTM for sequential processing.
    \item \textbf{MLFBK} \cite{MLFBK} utilizes BERT to incorporate explicit features and latent relations, enhancing prediction efficiency in knowledge tracing.
    \item \textbf{BiDKT} \cite{BiDKT} adapts BERT for knowledge tracing by leveraging bidirectional context in interaction histories, unlike traditional RNN-based models.
\end{itemize}
\subsubsection{Context-Aware Methods}
For context-aware methods, they utilize the context of questions to learn semantic knowledge. Particularly, we select the following four algorithms:
\begin{itemize}[leftmargin=*, align=left]
    \item \textbf{EERNN} \cite{EKT} combines student records and exercise content into a single vector, processed by a bidirectional LSTM, with two variants: EERNNM (Markov property) and EERNNA (Attention mechanism).
    \item \textbf{EKT} \cite{EKT} extends EERNN by using a knowledge state matrix, which captures the impact of exercises on multiple concepts, while a memory network tracks concept mastery.
    \item \textbf{RKT} \cite{RKT} uses relation-aware self-attention to integrate contextual information from exercises and performance data. It also includes a forgetting model with an exponentially decaying kernel to address interactions and forgetfulness.
    \item \textbf{DCL4KT-A} \cite{DCL4KT-A} introduces a difficulty-centered contrastive learning method and leverages LLMs to optimize and predict difficulty from unseen data.
\end{itemize}

\subsubsection{LLMs-based Methods}
Furthermore, LLM-based methods have good reasoning abilities with rich commonsense knowledge. We conduct four versions of LLMs based on both fine-tuning and prompting:
\begin{itemize}[leftmargin=*, align=left]
    \item \textbf{LLM-FT$_\mathrm{ID}$} finetunes the LLaMA model with instructions using QIDs and/or CIDs depending on the dataset.
    \item \textbf{LLM-FT$_\mathrm{TokenID}$} finetunes the LLaMA model by treating QID and CID as specific tokens, where their embeddings are updated during training.
    \item \textbf{LLM-FT$_\mathrm{Text}$} finetunes the LLaMA model using textual information of questions and concepts.
    \item \textbf{GPT-4o} inputs the same textual information as LLM-FT$_\mathrm{Text}$ directly into the GPT-4o framework.
\end{itemize}

For DL-based, PLMs-based, and Context-aware Methods, we only report the results from the original papers or other relevant experimental papers to ensure the reliability of the experimental results.
For LLMs-based Methods, since Assist2015 and Junyi have no textual information of the question and concept, we don't provide the results of LLM-FT$_{Text}$ and GPT-4o.
Note that we remove the information missed in the corresponding dataset (such as QID in Assist2015) from the prompt template in our experiments. 

\begin{table*}
\caption{Main results of our models and selected baselines. We give the results not reported by the original paper from \citet{MRT-KT,RKT,DKT}. Imp. means the relative improvement over the baseline LLM-FT$_\mathrm{ID}$. The best and suboptimal results are emphasized in \textbf{bold} and \underline{underline}.}
\label{table:main results.}
\vspace{-1mm}
\centering
\begin{tabular}{llcccccccc}
\hlineB{4}
& & \multicolumn{2}{c}{Assist2009} & \multicolumn{2}{c}{Assist2015} & \multicolumn{2}{c}{Junyi} & \multicolumn{2}{c}{Nips2020} \\ 
&  & AUC & ACC & AUC & ACC & AUC & ACC & AUC & ACC \\ \hline
\multirow{8}{*}{DL-based Methods}
&  DKT      & 0.7084 & 0.7221 & 0.7093 & 0.7542 & 0.8013 & 0.7200 & 0.7406 & 0.6878 \\
&  DKVMN    & 0.8157 & - & 0.7268 & - & 0.8027 & - & 0.7673 & 0.7016 \\
&  SAKT     & 0.8480 & - & 0.8540 & - & 0.8340 & 0.7570 & 0.7517 & 0.6879 \\
&  AKT      & 0.7767 & 0.7532 & 0.7211 & 0.7518 & \underline{0.8948} & 0.8215 & 0.7494 & 0.6930 \\
&  LPKT     & 0.7788 & 0.7325  & - & - & 0.7689  & \underline{0.8344}  & - & - \\
&  LBKT$^\dag$  & 0.7863 & 0.7380 & - & - & 0.7723  & \textbf{0.8362}  & - & - \\
&  AT-DKT   & 0.7574 & 0.7172 & - & - & 0.7581 & 0.8325 & 0.7816 & 0.7145 \\
&  MRT-KT   & 0.8223 & 0.7841 & - & - & - & - & - & -  \\ \hline
\multirow{3}{*}{PLMs-based Methods} &  BiDKT    & 0.7651 & - & 0.6766 & - & - & - & - & - \\
&  MLFBK    & \underline{0.8524} & - & - & - & - & - & - & - \\
&  LBKT$^\ddag$     & - & - & - & - & 0.8510 & 0.8320 & - & -\\ \hline
\multirow{4}{*}{Context-Aware Methods}
&  DCL4KT-A   & 0.8153 & - & - & - & - & - & - & - \\ 
&  EERNN   & - & - & - & - & 0.8370 & 0.7580 & - & - \\ 
&  EKT     & - & - & - & - & 0.8420 & 0.7590 & - & - \\ 
&  RKT     & - & - & - & - & 0.8600 & 0.7700 & - & - \\ \hline
\multirow{4}{*}{LLMs-based Methods} 
&  LLM-FT$_\mathrm{ID}$   & 0.8393 & 0.7592 & \underline{0.9092} & \underline{0.9092} & {0.8841} & 0.8071 & \underline{0.7890} & 0.6870 \\ 
&  LLM-FT$_\mathrm{TokenID}$   & 0.8143 & 0.7954 & 0.8386 & 0.8813 & 0.8663 & 0.8050 & 0.7774 & 0.5962 \\
&  LLM-FT$_\mathrm{Text}$   & 0.8407 & \underline{0.8119} & - & - & - & -  & 0.7762 & \underline{0.7211} \\ 
 &  GPT-4o   & - & 0.7274 & - & - & - & - & - & 0.6694 \\
\hline
\multirow{2}{*}{Ours} &  \texttt{\textbf{LLM-KT}}                 & \textbf{0.8870} & \textbf{0.8168} & \textbf{0.9356} & \textbf{0.9185} & \textbf{0.9018} & {0.8294} & \textbf{0.8291} & \textbf{0.7561} \\
&  \textcolor{blue}{Imp. (\%)}    & \textcolor{blue}{+5.68} & \textcolor{blue}{+7.59} & \textcolor{blue}{+2.90} & \textcolor{blue}{+1.02} & \textcolor{blue}{+2.00} & \textcolor{blue}{+2.76} & \textcolor{blue}{+5.08} & \textcolor{blue}{+10.06} \\ \hlineB{4}
\end{tabular}
\vspace{-2mm}
\end{table*}

\subsection{Implementation Details}
\label{sect:Implementation Details}
In our experiments, we use the deep learning framework PyTorch Lightning for its ease of use and efficient management of training processes. Following MRT-KT \cite{MRT-KT}, we divide the student dataset in a ratio of 8:1:1 for training, validation, and test sets. Then, we train on the training set, select the best model based on the validation set, and evaluate it on the test set. Our \textbf{\texttt{LLM-KT}} model is based on LLaMA2 \cite{llama2}, which is an advanced and commonly used open-source LLM with over 9,000 citations. We update our model using the parameter-efficient method LoRA, where the rank is $32$, alpha is $32$, and dropout is $0.1$. Each task is trained for a maximum of 10 epochs with a batch size of 32, using the gradient accumulation strategy. For the function $g$ used to merge the representations of context and sequence, we employ the addition operation. We utilize early stopping to avoid overfitting. Additionally, we use Adam as the optimizer with a learning rate of $3 \times 10^{-4}$ and a weight decay of $1 \times 10^{-5}$, utilizing a cosine learning rate scheduler. The sequence length of historical records is 100. We adopt LLaMA2-7B as the context encoder and AKT as the sequence encoder.

To elucidate the distinctions among diverse models within LLMs-based methods more perspicuously, we initially streamline the template in Section \ref{sect:methods} as follows.
\noindent\par
\noindent\resizebox{.99\columnwidth}{!}{
\begin{tcolorbox}[left=1mm,right=1mm,top=1mm,bottom=1mm]
\textbf{Input $x$.} \\
The student has previously, in chronological order, answered \underline{HistoryQues$_1$, HistoryQues$_2$,..., HistoryQues$_n$}. Please predict whether the student will answer \underline{TargetQues} correctly. Response with 'Yes' or 'No'. Response:
\end{tcolorbox}
}

For LLM-FT$_\mathrm{ID}$, we simply replace each ``HistoryQues'' in the template with the format ``question with ID=QID involving concept ID=CID correctly.'' At the same time, we change ``TargetQues'' to ``the next question with ID=QID involving concept ID=CID correctly.'' Note that the actual values of these IDs depend on the specific responses shared by students.
For LLM-FT$_\mathrm{TokenID}$, we adjust ``HistoryQues'' in the template with ``question with ID=[qid74] involving concept ID=[cid6] correctly'' and turn ``TargetQues'' into ``the next question with ID=[qid44] involving concept ID=[cid5] correctly.'' Here, [qid74]/[cid6]/[qid44]/[cid5] represents a newly introduced token that is finetuned on the target dataset to learn the semantic and interaction information in the given record. The number of newly added QID/CID tokens exactly matches the number of questions and concepts. 
As for LLM-FT$_\mathrm{Text}$ and GPT-4o, we replace ``HistoryQues'' and ``TargetQues'' in the template with specific textual questions. For example, we use ``Which symbol belongs in the box? Pic$_{749-0}$ A:$>$  B:$<$  C:$=$ D:$\ge$ Related knowledge concepts: Basic Arithmetic The student answered this question correctly.'' We then input these into LLaMA and GPT-4o, respectively.

\section{Experimental Analysis}
\label{sect:Experimental Analysis}
In this section, we conduct extensive experiments to evaluate the effectiveness of our \textbf{\texttt{LLM-KT}}. To be specific, we compare our model with the strong baselines in Section \ref{sect:Main Results}. Then, we explore the performance of the main components in Section \ref{sect:Ablation Studies} and investigate the influence of sequence length in Section \ref{sect:Influence of Sequence Length}. Finally, we analyze the influence of context encoder, sequence encoder, and ensemble function $g$ in Section \ref{sect:Further Analysis}.

\subsection{Main Results}
\label{sect:Main Results}
In this section, we report the results of our \textbf{\texttt{LLM-KT}} and the selected baselines across four benchmark datasets in terms of AUC and ACC (Table \ref{table:main results.}).  
To evaluate the effectiveness of our model, we compare it with four categories: DL-based methods, PLMs-based methods, context-aware methods, and LLMs-based methods.

From these results, we obtain the following observations. 
\textbf{First}, our proposed \textbf{\texttt{LLM-KT}} obtains the best performance in most cases. Particularly, we compare our model with four kinds of baselines, where DL-based methods focus on learning the sequence of IDs, PLMs/LLMs-based methods adopt PLMs/LLMs to improve the performance, and context-based methods integrate the textual information for KT. Our model outperforms all these baselines over three datasets, which indicates that it can integrate other modalities with LLMs effectively for knowledge tracing.
\textbf{Second}, \textbf{\texttt{LLM-KT}} captures the knowledge of the long context and interaction sequence effectively. Our proposed model outperforms the models that are fine-tuned using IDs and the original textual information (e.g., LLM-FT$_\mathrm{ID}$ and LLM-FT$_\mathrm{Text}$). 
\textbf{Third}, LLMs are not good at learning the sequence information or long context directly. For LLM$_\mathrm{TokenID}$, we regard the question ID as a specific token and fine-tune the LLMs to learn the representation of the question. We also input the long textual context of history records into GPT-4o. From the results, we find that they are not good at predicting the student's performance.


\begin{table}[t!]
\caption{The results of ablation studies.}
\label{table:ablation studies.}
\vspace{-1mm}
\centering
\begin{tabular}{lcccc}
\hlineB{4}
  & \multicolumn{2}{c}{Assist2009}  & \multicolumn{2}{c}{Nips2020} \\ 
   & AUC & ACC & AUC & ACC \\ \hline
LLM-KT                 & 0.8870 & 0.8168 & 0.8291 & 0.7561 \\ \hline
\multicolumn{5}{l}{\emph{Data Source}}     \\ \hline
- Question             & 0.8788 & 0.7937 & 0.7983 & 0.7439 \\
- Concept              & 0.8635 & 0.8053 & 0.8101 & 0.7276 \\ \hline
\multicolumn{5}{l}{\emph{Model Structure}}     \\ \hline
- Sequence             & 0.8616 & 0.8119 & 0.7958 & 0.7249 \\
- Context              & 0.8788 & 0.7937 & 0.8056 & 0.7358 \\
\hlineB{4}
\end{tabular}
\vspace{-2mm}
\end{table}

\subsection{Ablation Studies}
\label{sect:Ablation Studies}
To investigate the performance of the main components contained in our proposed \texttt{\textbf{LLM-KT}} (Table \ref{table:ablation studies.}). 
From the data source, we remove the ID and text of the question (- Question), and the ID and text of the concepts (- Concept) from our model. From the model structure, we remove the Plug-in Sequence (- Sequence) and Plug-in Context (- Context). 
Due to missing questions or concepts, we mainly conduct the experiments on the Assist2009 and Nips2020.

From the results, we observe that both question and concept information can help the model understand the student's state from the history record to improve the performance of knowledge tracing. For instance, removing questions from inputs will reduce 3.08 points in terms of AUC (0.8291 vs 0.7983). Additionally, our plug-in sequence and plug-in context effectively capture the sequence interaction behaviors and long textual context. Removing any one of them from our \texttt{\textbf{LLM-KT}} will reduce the performance. The textual context helps the model learn the complex semantic relationships between the questions and concepts, and the sequence information helps the model capture the interaction behaviors based on a sequence of IDs.

\begin{figure}[t!]
    \centering
    \begin{minipage}[b]{0.235\textwidth}
        \centering
        \includegraphics[width=\textwidth]{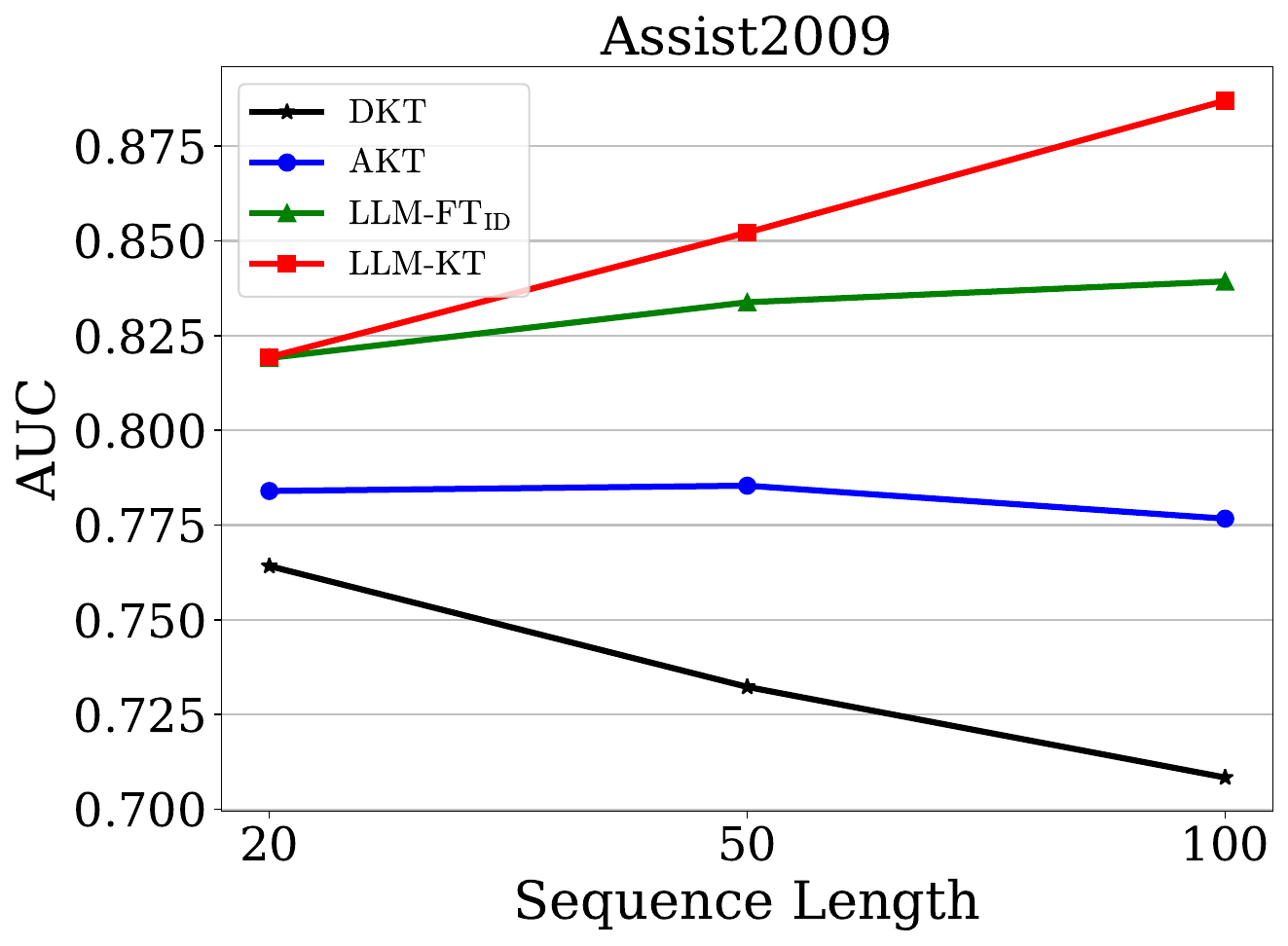}
    \end{minipage}
    \begin{minipage}[b]{0.235\textwidth}
        \centering
        \includegraphics[width=\textwidth]{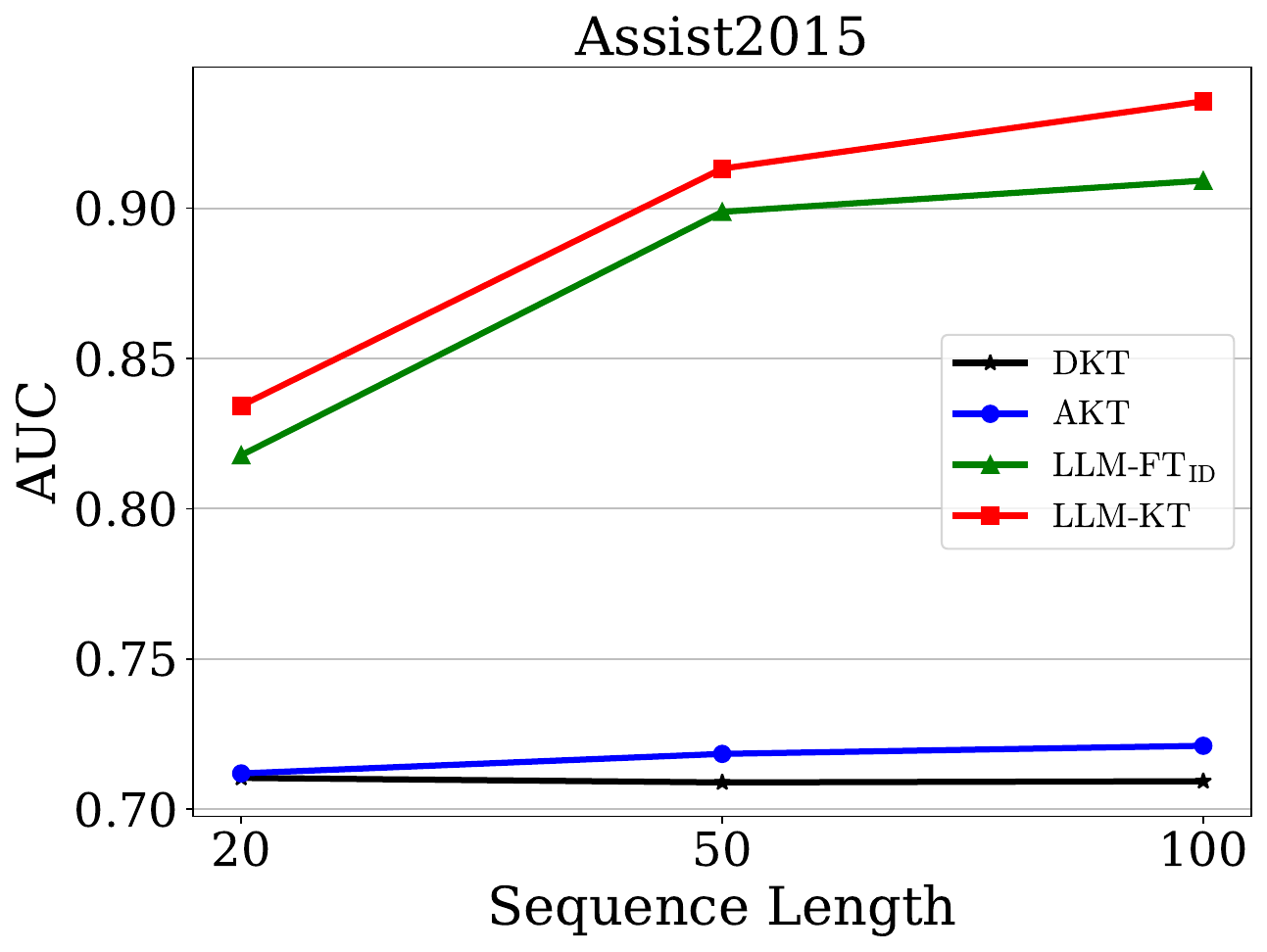}
    \end{minipage}
    \\
    \begin{minipage}[b]{0.235\textwidth}
        \centering
        \includegraphics[width=\textwidth]{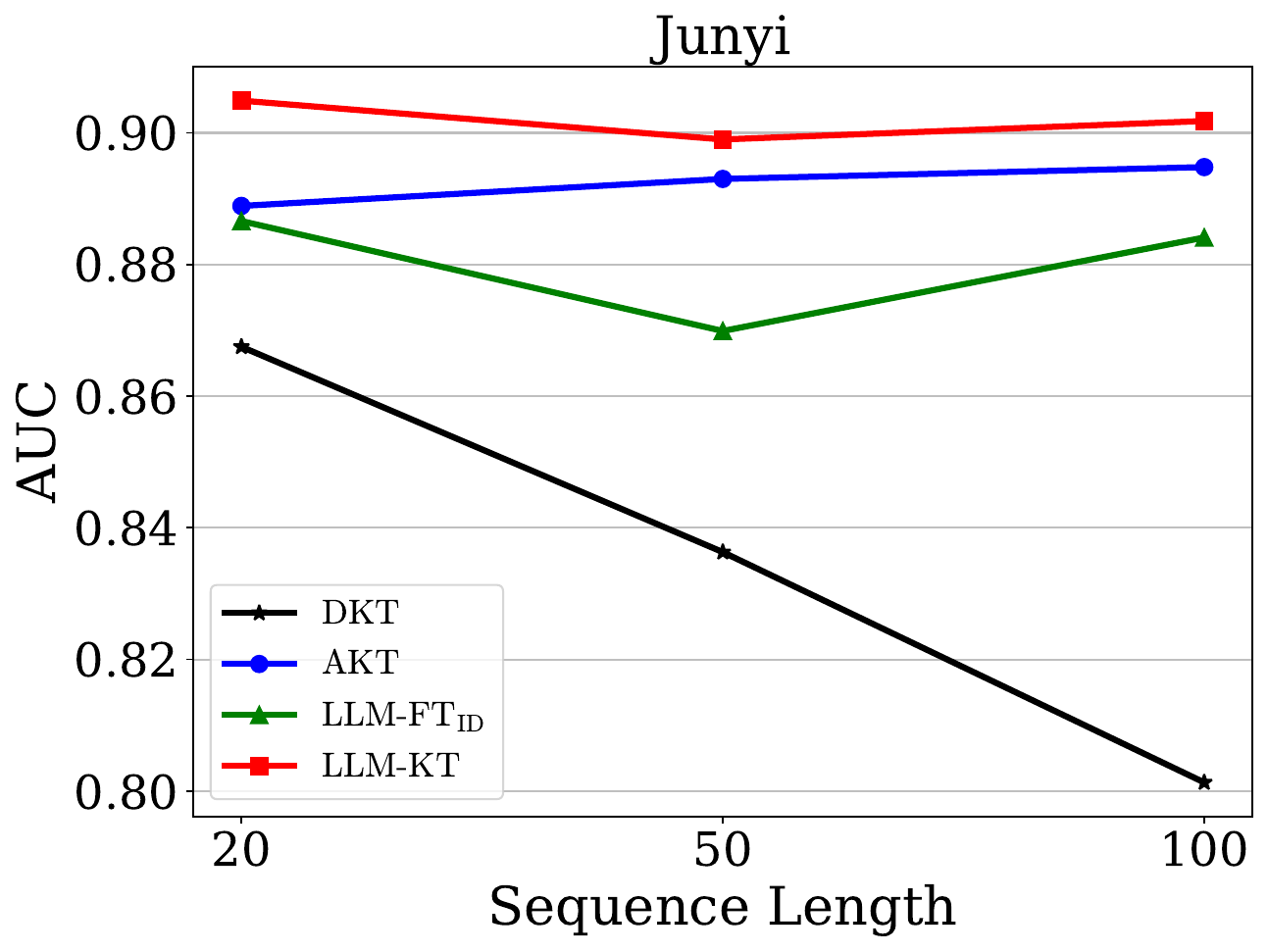}
    \end{minipage}
    \begin{minipage}[b]{0.235\textwidth}
        \centering
        \includegraphics[width=\textwidth]{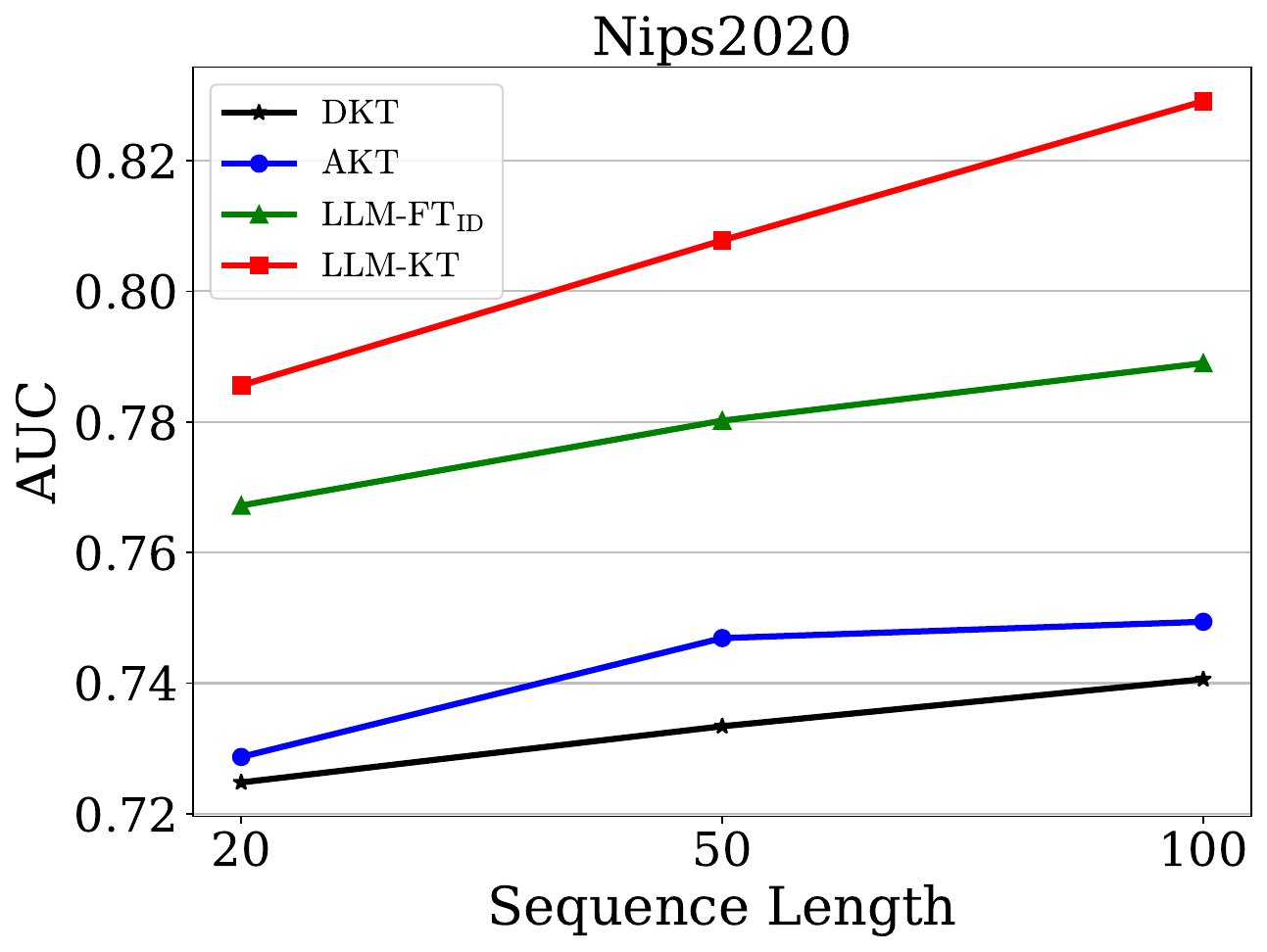}
    \end{minipage}
    \vspace{-2mm}
    \caption{Influence of sequence length on four different datasets in terms of AUC.}
    \label{fig:sequence length auc}
    \vspace{-1mm}
\end{figure}

\begin{figure}[t!]
    \centering
    \begin{minipage}[b]{0.235\textwidth}
        \centering
        \includegraphics[width=\textwidth]{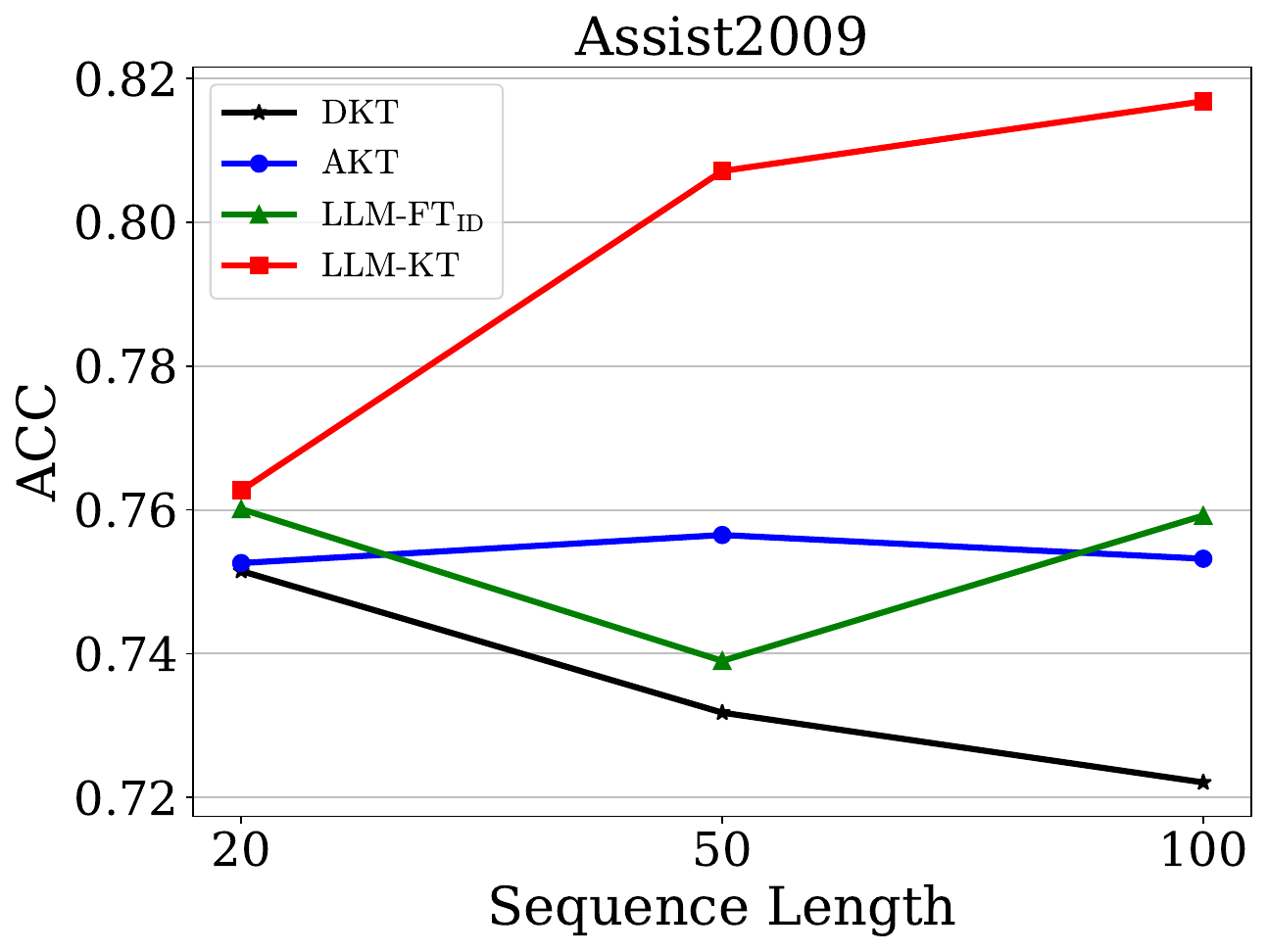}
    \end{minipage}
    \begin{minipage}[b]{0.235\textwidth}
        \centering
        \includegraphics[width=\textwidth]{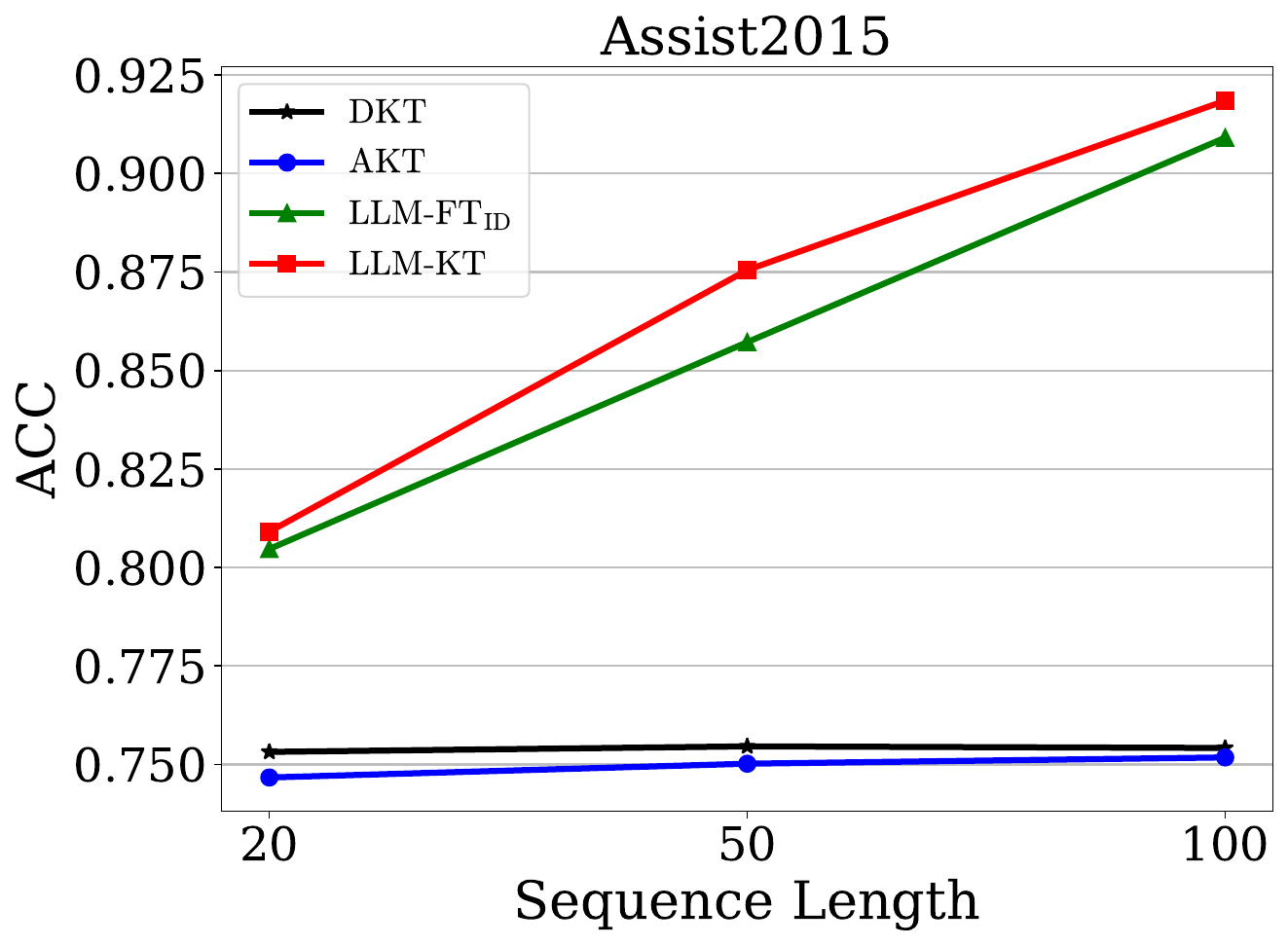}
    \end{minipage}
    \\
    \begin{minipage}[b]{0.235\textwidth}
        \centering
        \includegraphics[width=\textwidth]{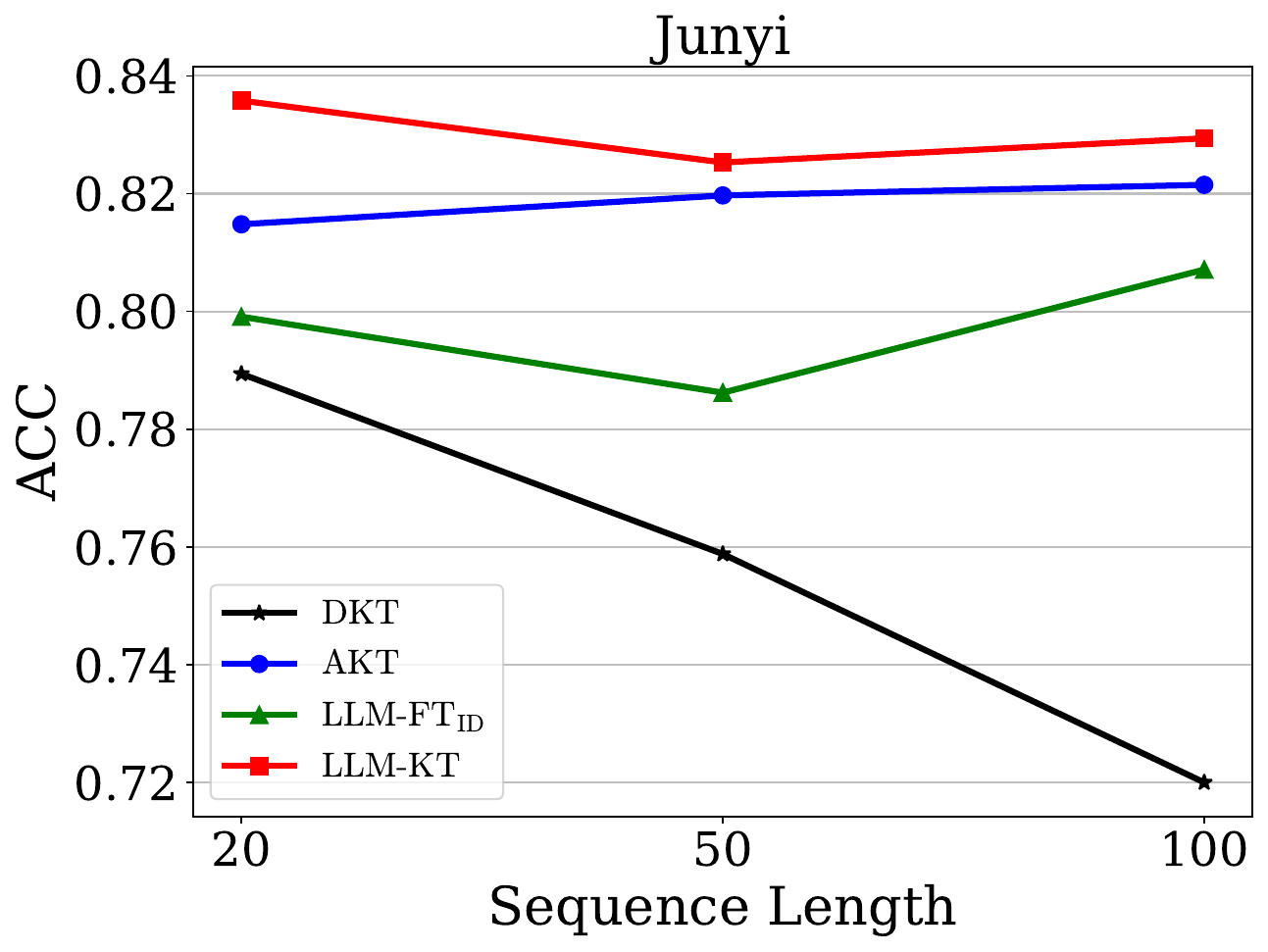}
    \end{minipage}
    \begin{minipage}[b]{0.235\textwidth}
        \centering
        \includegraphics[width=\textwidth]{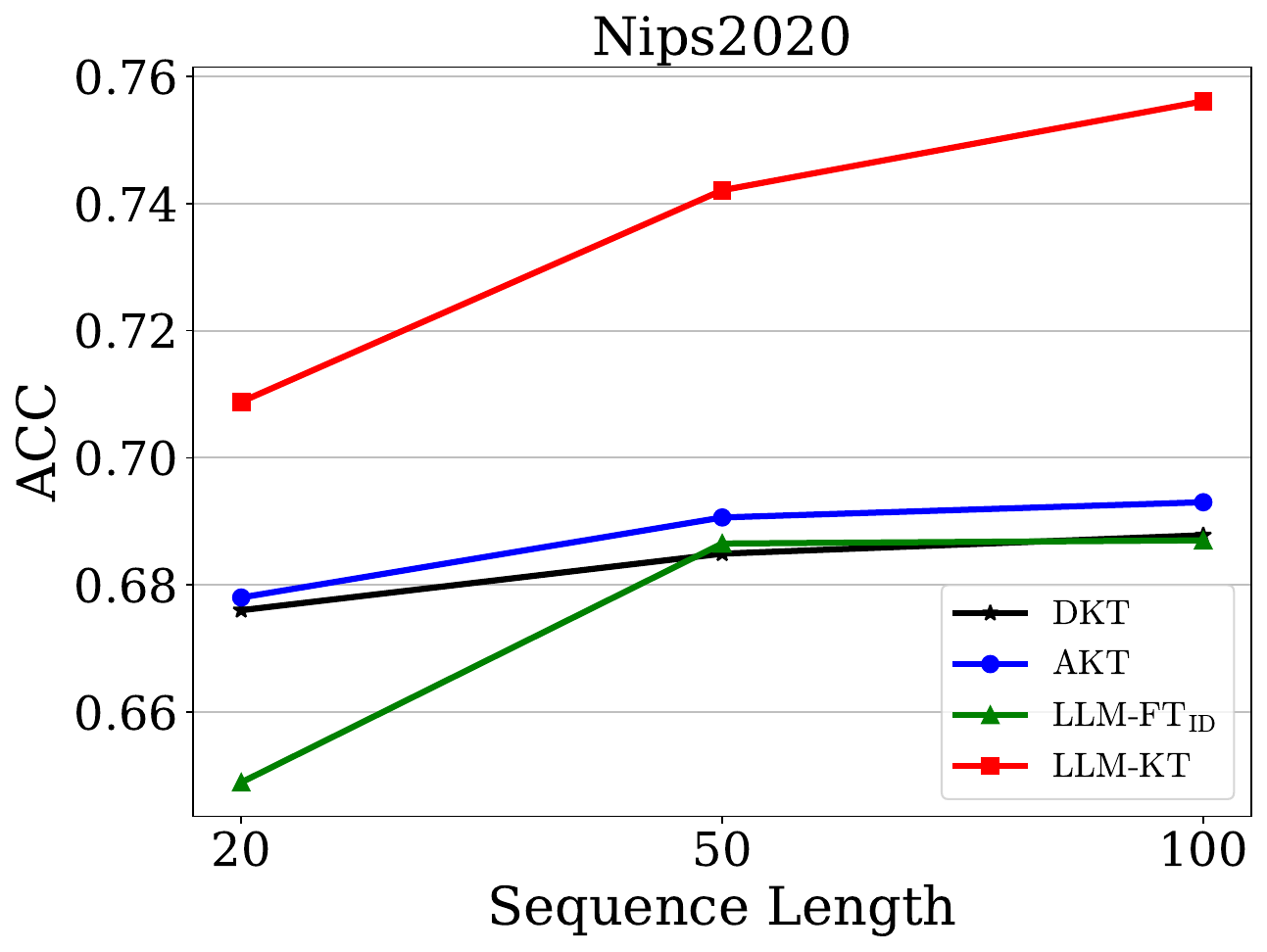}
    \end{minipage}
    \vspace{-1mm}
    \caption{Influence of sequence length on four different datasets in terms of ACC.}
    \label{fig:sequence length acc}
    \vspace{-2mm}
\end{figure}

\begin{table*}
\caption{Influence of sequence encoder. Average means the average score over four datasets.}
\label{table:sequence encoder}
\vspace{-1mm}
\centering
\begin{tabular}{lcccccccccc}
\hlineB{4}
    & \multicolumn{2}{c}{Assist2009} & \multicolumn{2}{c}{Assist2015} & \multicolumn{2}{c}{Junyi} & \multicolumn{2}{c}{Nips2020} & \multicolumn{2}{c}{Average} \\
    & AUC & ACC & AUC & ACC & AUC & ACC & AUC & ACC & AUC & ACC      \\
    \hline
LLM-FT$_\text{TokenID}$ & 0.8143 & {0.7954} & 0.8386 & 0.8813 & 0.8663 & 0.8050 & 0.7774 & 0.5962  & 0.8242  & 0.7695\\
DKT & 0.8759 & \textbf{0.8218} & 0.9350  & \textbf{0.9204} & \textbf{0.9027} & \textbf{0.8303} & 0.8149 & 0.7453   & 0.8821 & 0.8295 \\
AKT & \textbf{0.8870} & 0.8168 & \textbf{0.9356}  & 0.9185 & 0.9018 & 0.8294 & \textbf{0.8291} & \textbf{0.7561}   & \textbf{0.8884} & \textbf{0.8302} \\
\hlineB{4}
\end{tabular}
\end{table*}

\subsection{Influence of Sequence Length} 
\label{sect:Influence of Sequence Length}
In this section, we examine how sequence length affects knowledge tracing.
We present our model's results alongside several robust baselines, measured by AUC and ACC (Figure \ref{fig:sequence length auc} and \ref{fig:sequence length acc}). 
We define the sequence lengths as 20, 50, or 100.

From the results, we find that our model outperforms the other methods across all sequence lengths, which indicates that \texttt{\textbf{LLM-KT}} can capture the students' states with long text and ID sequences effectively. Specifically, our model improves by more than 4 points of AUC compared to DKT and AKT on Nips2020.
With the increase in sequence length, our model can effectively improve performance in most cases, while the improvements in previous studies are limited and may even decline. For example, \texttt{\textbf{LLM-KT}} with a sequence length of 100 outperforms the one with 20 questions by about 10 points in terms of AUC over Assist2015.
The Junyi dataset primarily focuses on short-term records, and extending the sequence length may reduce performance due to unrelated questions in the history record.
Though our model experiences a performance drop when increasing the length, the decrease is subtle. 
For DKT, it is an LSTM-based model that is not effective at capturing long sequences. 
We would like to explore how to reduce the influence of unrelated information in lengthy sequences in the future.

\begin{table}[t!]
\caption{Influence of context encoder.}
\label{table:context encoder}
\vspace{-1mm}
\centering
\setlength{\tabcolsep}{1.3mm}{
\begin{tabular}{ccccccc}
\hlineB{4}
\multicolumn{1}{l}{} & \multicolumn{2}{c}{Assist2009} & \multicolumn{2}{c}{Nips2020} & \multicolumn{2}{c}{Average} \\ \hline
\multicolumn{1}{l}{} & AUC & ACC & AUC & ACC  & AUC & ACC       \\ \hline
BERT         & 0.8770         & 0.7987        & 0.8116      & 0.7331    & 0.8443  &  0.7659 \\ 
MPNET       & 0.8749         & 0.8135        & 0.8231      & \textbf{0.7561}   & 0.8490 & 0.7848 \\ 
LLaMA2       & \textbf{0.8870}         & \textbf{0.8168}        & \textbf{0.8291}      & \textbf{0.7561}  & \textbf{0.8581} & \textbf{0.7865} \\ \hlineB{4}
\end{tabular}}
\vspace{-1mm}
\end{table}

\subsection{Further Analysis}
\label{sect:Further Analysis}
\subsubsection{Influence of Context Encoder}
We investigate the influence of different context encoders and provide their average performance for both AUC and ACC (Table \ref{table:context encoder}). 
For the context encoder, we select three typical sentence encoders: LLaMA2\footnote{\url{https://huggingface.co/meta-llama/Llama-2-7b-chat-hf}}, BERT\footnote{\url{https://huggingface.co/google-bert/bert-base-uncased}}, and all-mpnet-base-v2 (MPNET)\footnote{\url{https://huggingface.co/sentence-transformers/all-mpnet-base-v2}}. 
We input the entire question and concepts into the Context-Encoder to generate a vector $r_{QText}$ and $r_{CText}$. Consequently, we derive a single vector for each question and all associated concepts, which remains unaffected by increases in question length or the number of concepts. Specifically, the vectors are derived by averaging the hidden states from the last layer along the sequence length for the BERT and LLaMA2 models, respectively. For the MPNET model, we calculate the the vector using SentenceTransformer. The dimensions $d^t$ are 768, 768, and 4096 for BERT, MPNET, and the LLaMA2-based Context-Encoder, respectively. We develop a Context-Adapter to convert $d^t$ into the embedding layer dimension $d^e$ of LLaMA2 (e.g., 4096).

Among the models, the LLaMA2-enhanced model achieves the highest average AUC (0.8581) and ACC (0.7865), demonstrating its superiority in capturing long context and making accurate predictions. 
In contrast, the corresponding model with BERT shows the lowest overall performance (AUC: 0.8443, ACC: 0.7659), particularly struggling with the Nips2020 dataset where it achieves the lowest AUC (0.8116) and ACC (0.7331). 
Particularly, LLaMA2 achieves a 2-point increase in ACC on average when compared to BERT (0.7865 vs 0.7659).
By training on a large-scale corpus with huge parameters, LLaMA2 learns powerful text embeddings, which help LLMs acquire the long textual knowledge of the questions. 
For MPNET, it is fine-tuned on 1B sentence pairs using a contrastive learning objective to improve the sentence representation. 
Based on the experiments, we recommend using MPNET, which is the same size as BERT and achieves competitive performance to LLaMA2.

\subsubsection{Influence of Sequence Encoder}
We investigate the influence of different sequence encoders and compare them with LLM-FT$_\text{TokenID}$ over four datasets (Table \ref{table:sequence encoder}).  
Particularly, we employ two sequence encoders: DKT \cite{DKT} and AKT \cite{AKT} to model the sequence of IDs.  
For LLM-FT$_\text{TokenID}$, it initiates the embedding of the question- and concept-specific token randomly and updates the embeddings by fine-tuning on the training dataset.  
Unlike LLM-FT$_\text{TokenID}$, we use representations of question IDs and concept IDs learned by DKT and AKT to initiate the tokens' embeddings.  

The findings indicate that models utilizing DKT and AKT perform better than LLM-FT$_\text{TokenID}$, showcasing their robust ability to model knowledge tracing tasks effectively. 
The AKT-based model achieves over a 15-point enhancement in terms of ACC over Nips2020 when compared to LLM-FT$_\text{TokenID}$. 
It suggests that the embeddings derived from traditional sequence models capture extensive semantic and interaction knowledge. 
Also, LLMs cannot capture the knowledge well by simply fine-tuning on sequence of IDs.
Additionally, DKT and AKT obtain comparable results over these four datasets (0.8295 vs 0.8302 for average ACC).

\begin{table}[t!]
\caption{Influence of function $g$.}
\label{table:merge methods}
\vspace{-1mm}
\centering
\setlength{\tabcolsep}{1.3mm}{
\begin{tabular}{clcccccc}
\hlineB{4}
\multicolumn{2}{c}{}                      & \multicolumn{2}{c}{Assist2009} & \multicolumn{2}{c}{Nips2020}  & \multicolumn{2}{c}{Average} \\
\multicolumn{2}{c}{}                      & AUC & ACC & AUC & ACC  & AUC & ACC      \\ \hline
\multicolumn{2}{c}{Concat}    & 0.8851         & 0.8102        & 0.8168      & 0.7466   & 0.8510 & 0.7784  \\ 
\multicolumn{2}{c}{Avg}       & 0.8763         & \textbf{0.8185}        & 0.8262      & 0.7466    & 0.8513 &  0.7826  \\ 
\multicolumn{2}{c}{Add}       & \textbf{0.8870}         & 0.8168        & \textbf{0.8291}      & \textbf{0.7561}   &  \textbf{0.8581} &   \textbf{0.7865} \\ \hlineB{4}
\end{tabular}}
\vspace{-1mm}
\end{table}

\subsubsection{Influence of Function $g$}
To combine the representations of context and sequence, we use a function $g$ to merge them, as mentioned in Equation \ref{equ:combine}. 
In our experiment, we explore the influence of different methods on two datasets, Assist2009 and Nips2020, using AUC and ACC as performance metrics (Table \ref{table:merge methods}).
To be specific, we evaluate the performance of three strategies, including concatenation (Concat), average (Avg), and addition (Add). 
For the Assist2009 dataset, the Add strategy achieves the highest AUC (0.8870) and ties for the best ACC (0.8168) with the Concat strategy. In contrast, for the Nips2020 dataset, the Add strategy again outperforms others in both AUC (0.8291) and ACC (0.7561). When averaging the performance across both datasets, the Add strategy consistently demonstrates superior results, achieving the highest overall AUC (0.8581) and ACC (0.7865). These findings indicate that the Add strategy is the most effective method for combining features across the evaluated scenarios.

\section{Conclusions and Further Work}
\label{sec:Conclusions and Further Work}
In this paper, we propose a large language models-based framework for knowledge tracing (\texttt{\textbf{LLM-KT}}), leveraging the Plug-and-Play Instruction to elegantly translate the sequential tracing task into a language modeling problem. This approach incorporates multiple modalities, such as the textual information of the questions and the interaction behavior information from traditional sequence models, into the large model. Extensive experiments demonstrate that our method achieves superior performance, surpassing previous state-of-the-art results. Ablation studies confirm the effectiveness of injecting both textual information and sequence behavior information. Additionally, we analyze the impact of sequence length and the roles of the context or sequence encoders.
In the future, we would like to explore alternative integration methods and alignment strategies. It is also interesting to design a more effective algorithm to capture the long-term sequence and reduce the impact of unrelated information.


\bibliographystyle{ACM-Reference-Format}
\bibliography{sample-base}


\begin{thebibliography}{39}


\ifx \showCODEN    \undefined \def \showCODEN     #1{\unskip}     \fi
\ifx \showDOI      \undefined \def \showDOI       #1{#1}\fi
\ifx \showISBNx    \undefined \def \showISBNx     #1{\unskip}     \fi
\ifx \showISBNxiii \undefined \def \showISBNxiii  #1{\unskip}     \fi
\ifx \showISSN     \undefined \def \showISSN      #1{\unskip}     \fi
\ifx \showLCCN     \undefined \def \showLCCN      #1{\unskip}     \fi
\ifx \shownote     \undefined \def \shownote      #1{#1}          \fi
\ifx \showarticletitle \undefined \def \showarticletitle #1{#1}   \fi
\ifx \showURL      \undefined \def \showURL       {\relax}        \fi
\providecommand\bibfield[2]{#2}
\providecommand\bibinfo[2]{#2}
\providecommand\natexlab[1]{#1}
\providecommand\showeprint[2][]{arXiv:#2}

\bibitem[Abdelrahman et~al\mbox{.}(2023)]%
        {abdelrahman2023knowledge}
\bibfield{author}{\bibinfo{person}{Ghodai Abdelrahman}, \bibinfo{person}{Qing Wang}, {and} \bibinfo{person}{Bernardo Nunes}.} \bibinfo{year}{2023}\natexlab{}.
\newblock \showarticletitle{Knowledge tracing: A survey}.
\newblock \bibinfo{journal}{\emph{Comput. Surveys}} \bibinfo{volume}{55}, \bibinfo{number}{11} (\bibinfo{year}{2023}), \bibinfo{pages}{1--37}.
\newblock


\bibitem[bigdata ustc(2021)]%
        {Edudata}
\bibfield{author}{\bibinfo{person}{bigdata ustc}.} \bibinfo{year}{2021}\natexlab{}.
\newblock \bibinfo{title}{EduData}.
\newblock \bibinfo{howpublished}{\url{https://github.com/bigdata-ustc/EduData}}.
\newblock


\bibitem[Cheng et~al\mbox{.}(2022)]%
        {AdaptKT}
\bibfield{author}{\bibinfo{person}{Song Cheng}, \bibinfo{person}{Qi Liu}, \bibinfo{person}{Enhong Chen}, \bibinfo{person}{Kai Zhang}, \bibinfo{person}{Zhenya Huang}, \bibinfo{person}{Yu Yin}, \bibinfo{person}{Xiaoqing Huang}, {and} \bibinfo{person}{Yu Su}.} \bibinfo{year}{2022}\natexlab{}.
\newblock \showarticletitle{AdaptKT: A domain adaptable method for knowledge tracing}. In \bibinfo{booktitle}{\emph{Proceedings of the Fifteenth ACM International Conference on Web Search and Data Mining}}. \bibinfo{pages}{123--131}.
\newblock


\bibitem[Corbett and Anderson(1995)]%
        {BKT}
\bibfield{author}{\bibinfo{person}{Albert~T. Corbett} {and} \bibinfo{person}{John~R. Anderson}.} \bibinfo{year}{1995}\natexlab{}.
\newblock \showarticletitle{Knowledge tracing: Modeling the acquisition of procedural knowledge}.
\newblock \bibinfo{journal}{\emph{User Modelling and User-Adapted Interaction}} (\bibinfo{date}{Jan} \bibinfo{year}{1995}), \bibinfo{pages}{253–278}.
\newblock
\urldef\tempurl%
\url{https://doi.org/10.1007/bf01099821}
\showDOI{\tempurl}


\bibitem[Cui et~al\mbox{.}(2024)]%
        {DGEKT}
\bibfield{author}{\bibinfo{person}{Chaoran Cui}, \bibinfo{person}{Yumo Yao}, \bibinfo{person}{Chunyun Zhang}, \bibinfo{person}{Hebo Ma}, \bibinfo{person}{Yuling Ma}, \bibinfo{person}{Zhaochun Ren}, \bibinfo{person}{Chen Zhang}, {and} \bibinfo{person}{James Ko}.} \bibinfo{year}{2024}\natexlab{}.
\newblock \showarticletitle{DGEKT: A Dual Graph Ensemble Learning Method for Knowledge Tracing}.
\newblock \bibinfo{journal}{\emph{ACM Trans. Inf. Syst.}} \bibinfo{volume}{42}, \bibinfo{number}{3}, Article \bibinfo{articleno}{78} (\bibinfo{date}{Jan.} \bibinfo{year}{2024}), \bibinfo{numpages}{24}~pages.
\newblock
\showISSN{1046-8188}
\urldef\tempurl%
\url{https://doi.org/10.1145/3638350}
\showDOI{\tempurl}


\bibitem[Cui et~al\mbox{.}(2023)]%
        {MRT-KT}
\bibfield{author}{\bibinfo{person}{Jiajun Cui}, \bibinfo{person}{Zeyuan Chen}, \bibinfo{person}{Aimin Zhou}, \bibinfo{person}{Jianyong Wang}, {and} \bibinfo{person}{Wei Zhang}.} \bibinfo{year}{2023}\natexlab{}.
\newblock \showarticletitle{Fine-grained interaction modeling with multi-relational transformer for knowledge tracing}.
\newblock \bibinfo{journal}{\emph{ACM Transactions on Information Systems}} \bibinfo{volume}{41}, \bibinfo{number}{4} (\bibinfo{year}{2023}), \bibinfo{pages}{1--26}.
\newblock


\bibitem[Devlin et~al\mbox{.}(2019)]%
        {bert}
\bibfield{author}{\bibinfo{person}{Jacob Devlin}, \bibinfo{person}{Ming-Wei Chang}, \bibinfo{person}{Kenton Lee}, {and} \bibinfo{person}{Kristina Toutanova}.} \bibinfo{year}{2019}\natexlab{}.
\newblock \showarticletitle{BERT: Pre-training of Deep Bidirectional Transformers for Language Understanding}. In \bibinfo{booktitle}{\emph{North American Chapter of the Association for Computational Linguistics}}.
\newblock
\urldef\tempurl%
\url{https://api.semanticscholar.org/CorpusID:52967399}
\showURL{%
\tempurl}


\bibitem[Elman(1990)]%
        {RNN}
\bibfield{author}{\bibinfo{person}{Jeffrey~L Elman}.} \bibinfo{year}{1990}\natexlab{}.
\newblock \showarticletitle{Finding structure in time}.
\newblock \bibinfo{journal}{\emph{Cognitive science}} \bibinfo{volume}{14}, \bibinfo{number}{2} (\bibinfo{year}{1990}), \bibinfo{pages}{179--211}.
\newblock


\bibitem[Ghosh et~al\mbox{.}(2020)]%
        {AKT}
\bibfield{author}{\bibinfo{person}{Aritra Ghosh}, \bibinfo{person}{Neil Heffernan}, {and} \bibinfo{person}{Andrew~S Lan}.} \bibinfo{year}{2020}\natexlab{}.
\newblock \showarticletitle{Context-aware attentive knowledge tracing}. In \bibinfo{booktitle}{\emph{Proceedings of the 26th ACM SIGKDD international conference on knowledge discovery \& data mining}}. \bibinfo{pages}{2330--2339}.
\newblock


\bibitem[Green({[n.\,d.]})]%
        {IRT}
\bibfield{author}{\bibinfo{person}{Bert~F. Green}.} \bibinfo{year}{[n.\,d.]}\natexlab{}.
\newblock \showarticletitle{A general solution for the latent class model of latent structure analysis}.
\newblock \bibinfo{journal}{\emph{Psychometrika}} \bibinfo{volume}{16}, \bibinfo{number}{2} (\bibinfo{year}{[n.\,d.]}), \bibinfo{pages}{151–166}.
\newblock
\urldef\tempurl%
\url{https://doi.org/10.1007/bf02289112}
\showDOI{\tempurl}


\bibitem[Hochreiter(1997)]%
        {hochreiter1997long}
\bibfield{author}{\bibinfo{person}{S Hochreiter}.} \bibinfo{year}{1997}\natexlab{}.
\newblock \showarticletitle{Long Short-term Memory}.
\newblock \bibinfo{journal}{\emph{Neural Computation MIT-Press}} (\bibinfo{year}{1997}).
\newblock


\bibitem[J. et~al\mbox{.}(2021)]%
        {LoRA}
\bibfield{author}{\bibinfo{person}{HuEdward J.}, \bibinfo{person}{Yulong Shen}, \bibinfo{person}{Phillip Wallis}, \bibinfo{person}{Zeyuan Allen-Zhu}, \bibinfo{person}{Yuanzhi Li}, \bibinfo{person}{Shean Wang}, {and} \bibinfo{person}{Weizhu Chen}.} \bibinfo{year}{2021}\natexlab{}.
\newblock \showarticletitle{LoRA: Low-Rank Adaptation of Large Language Models.}
\newblock \bibinfo{journal}{\emph{arXiv: Computation and Language,arXiv: Computation and Language}} (\bibinfo{date}{Jun} \bibinfo{year}{2021}).
\newblock


\bibitem[Lee et~al\mbox{.}(2024)]%
        {DCL4KT-A}
\bibfield{author}{\bibinfo{person}{Unggi Lee}, \bibinfo{person}{Sungjun Yoon}, \bibinfo{person}{Joon~Seo Yun}, \bibinfo{person}{Kyoungsoo Park}, \bibinfo{person}{Younghoon Jung}, \bibinfo{person}{Damji Stratton}, {and} \bibinfo{person}{Hyeoncheol Kim}.} \bibinfo{year}{2024}\natexlab{}.
\newblock \showarticletitle{Difficulty-Focused Contrastive Learning for Knowledge Tracing with a Large Language Model-Based Difficulty Prediction}. In \bibinfo{booktitle}{\emph{Proceedings of the 2024 Joint International Conference on Computational Linguistics, Language Resources and Evaluation, {LREC/COLING} 2024, 20-25 May, 2024, Torino, Italy}}, \bibfield{editor}{\bibinfo{person}{Nicoletta Calzolari}, \bibinfo{person}{Min{-}Yen Kan}, \bibinfo{person}{V{\'{e}}ronique Hoste}, \bibinfo{person}{Alessandro Lenci}, \bibinfo{person}{Sakriani Sakti}, {and} \bibinfo{person}{Nianwen Xue}} (Eds.). \bibinfo{publisher}{{ELRA} and {ICCL}}, \bibinfo{pages}{4891--4900}.
\newblock
\urldef\tempurl%
\url{https://aclanthology.org/2024.lrec-main.438}
\showURL{%
\tempurl}


\bibitem[Li et~al\mbox{.}(2023)]%
        {MLFBK}
\bibfield{author}{\bibinfo{person}{Zhaoxing Li}, \bibinfo{person}{Mark Jacobsen}, \bibinfo{person}{Lei Shi}, \bibinfo{person}{Yunzhan Zhou}, {and} \bibinfo{person}{Jindi Wang}.} \bibinfo{year}{2023}\natexlab{}.
\newblock \showarticletitle{Broader and Deeper: A Multi-Features with Latent Relations BERT Knowledge Tracing Model}. In \bibinfo{booktitle}{\emph{Responsive and Sustainable Educational Futures}}, \bibfield{editor}{\bibinfo{person}{Olga Viberg}, \bibinfo{person}{Ioana Jivet}, \bibinfo{person}{Pedro J. Mu{\~{n}}oz-Merino}, \bibinfo{person}{Maria Perifanou}, {and} \bibinfo{person}{Tina Papathoma}} (Eds.). \bibinfo{publisher}{Springer Nature Switzerland}, \bibinfo{address}{Cham}, \bibinfo{pages}{183--197}.
\newblock
\showISBNx{978-3-031-42682-7}


\bibitem[Li et~al\mbox{.}(2024)]%
        {lstm_bert}
\bibfield{author}{\bibinfo{person}{Zhaoxing Li}, \bibinfo{person}{Jujie Yang}, \bibinfo{person}{Jindi Wang}, \bibinfo{person}{Lei Shi}, {and} \bibinfo{person}{Sebastian Stein}.} \bibinfo{year}{2024}\natexlab{}.
\newblock \showarticletitle{Integrating lstm and bert for long-sequence data analysis in intelligent tutoring systems}.
\newblock \bibinfo{journal}{\emph{arXiv preprint arXiv:2405.05136}} (\bibinfo{year}{2024}).
\newblock


\bibitem[Liu et~al\mbox{.}(2019a)]%
        {EKT}
\bibfield{author}{\bibinfo{person}{Qi Liu}, \bibinfo{person}{Zhenya Huang}, \bibinfo{person}{Yu Yin}, \bibinfo{person}{Enhong Chen}, \bibinfo{person}{Hui Xiong}, \bibinfo{person}{Yu Su}, {and} \bibinfo{person}{Guoping Hu}.} \bibinfo{year}{2019}\natexlab{a}.
\newblock \showarticletitle{Ekt: Exercise-aware knowledge tracing for student performance prediction}.
\newblock \bibinfo{journal}{\emph{IEEE Transactions on Knowledge and Data Engineering}} \bibinfo{volume}{33}, \bibinfo{number}{1} (\bibinfo{year}{2019}), \bibinfo{pages}{100--115}.
\newblock


\bibitem[Liu et~al\mbox{.}(2019b)]%
        {RoBERT}
\bibfield{author}{\bibinfo{person}{Yinhan Liu}, \bibinfo{person}{Myle Ott}, \bibinfo{person}{Naman Goyal}, \bibinfo{person}{Jingfei Du}, \bibinfo{person}{Mandar Joshi}, \bibinfo{person}{Danqi Chen}, \bibinfo{person}{Omer Levy}, \bibinfo{person}{Mike Lewis}, \bibinfo{person}{Luke Zettlemoyer}, {and} \bibinfo{person}{Veselin Stoyanov}.} \bibinfo{year}{2019}\natexlab{b}.
\newblock \showarticletitle{Roberta: A robustly optimized bert pretraining approach}.
\newblock \bibinfo{journal}{\emph{arXiv preprint arXiv:1907.11692}} (\bibinfo{year}{2019}).
\newblock


\bibitem[Liu et~al\mbox{.}(2021)]%
        {PEBG}
\bibfield{author}{\bibinfo{person}{Yunfei Liu}, \bibinfo{person}{Yang Yang}, \bibinfo{person}{Xianyu Chen}, \bibinfo{person}{Jian Shen}, \bibinfo{person}{Haifeng Zhang}, {and} \bibinfo{person}{Yong Yu}.} \bibinfo{year}{2021}\natexlab{}.
\newblock \showarticletitle{Improving knowledge tracing via pre-training question embeddings}. In \bibinfo{booktitle}{\emph{Proceedings of the Twenty-Ninth International Conference on International Joint Conferences on Artificial Intelligence}}. \bibinfo{pages}{1577--1583}.
\newblock


\bibitem[Nagatani et~al\mbox{.}(2019)]%
        {DKT+Forget}
\bibfield{author}{\bibinfo{person}{Koki Nagatani}, \bibinfo{person}{Qian Zhang}, \bibinfo{person}{Masahiro Sato}, \bibinfo{person}{Yan-Ying Chen}, \bibinfo{person}{Francine Chen}, {and} \bibinfo{person}{Tomoko Ohkuma}.} \bibinfo{year}{2019}\natexlab{}.
\newblock \showarticletitle{Augmenting knowledge tracing by considering forgetting behavior}. In \bibinfo{booktitle}{\emph{The world wide web conference}}. \bibinfo{pages}{3101--3107}.
\newblock


\bibitem[Nakagawa et~al\mbox{.}(2019)]%
        {GKT}
\bibfield{author}{\bibinfo{person}{Hiromi Nakagawa}, \bibinfo{person}{Yusuke Iwasawa}, {and} \bibinfo{person}{Yutaka Matsuo}.} \bibinfo{year}{2019}\natexlab{}.
\newblock \showarticletitle{Graph-based Knowledge Tracing: Modeling Student Proficiency Using Graph Neural Network}. In \bibinfo{booktitle}{\emph{IEEE/WIC/ACM International Conference on Web Intelligence}}.
\newblock
\urldef\tempurl%
\url{https://doi.org/10.1145/3350546.3352513}
\showDOI{\tempurl}


\bibitem[Pandey and Karypis(2019)]%
        {SAKT}
\bibfield{author}{\bibinfo{person}{Shalini Pandey} {and} \bibinfo{person}{George Karypis}.} \bibinfo{year}{2019}\natexlab{}.
\newblock \showarticletitle{A self-attentive model for knowledge tracing}.
\newblock \bibinfo{journal}{\emph{arXiv preprint arXiv:1907.06837}} (\bibinfo{year}{2019}).
\newblock


\bibitem[Pandey and Srivastava(2020)]%
        {RKT}
\bibfield{author}{\bibinfo{person}{Shalini Pandey} {and} \bibinfo{person}{Jaideep Srivastava}.} \bibinfo{year}{2020}\natexlab{}.
\newblock \showarticletitle{RKT: Relation-Aware Self-Attention for Knowledge Tracing}.
\newblock \bibinfo{journal}{\emph{Proceedings of the 29th ACM International Conference on Information \& Knowledge Management}} (\bibinfo{year}{2020}).
\newblock
\urldef\tempurl%
\url{https://api.semanticscholar.org/CorpusID:221370579}
\showURL{%
\tempurl}


\bibitem[Piech et~al\mbox{.}(2015)]%
        {DKT}
\bibfield{author}{\bibinfo{person}{Chris Piech}, \bibinfo{person}{Jonathan Bassen}, \bibinfo{person}{Jonathan Huang}, \bibinfo{person}{Surya Ganguli}, \bibinfo{person}{Mehran Sahami}, \bibinfo{person}{Leonidas~J Guibas}, {and} \bibinfo{person}{Jascha Sohl-Dickstein}.} \bibinfo{year}{2015}\natexlab{}.
\newblock \showarticletitle{Deep knowledge tracing}.
\newblock \bibinfo{journal}{\emph{Advances in neural information processing systems}}  \bibinfo{volume}{28} (\bibinfo{year}{2015}).
\newblock


\bibitem[Shen et~al\mbox{.}(2021)]%
        {LPKT}
\bibfield{author}{\bibinfo{person}{Shuanghong Shen}, \bibinfo{person}{Qi Liu}, \bibinfo{person}{Enhong Chen}, \bibinfo{person}{Zhenya Huang}, \bibinfo{person}{Wei Huang}, \bibinfo{person}{Yu Yin}, \bibinfo{person}{Yu Su}, {and} \bibinfo{person}{Shijin Wang}.} \bibinfo{year}{2021}\natexlab{}.
\newblock \showarticletitle{Learning process-consistent knowledge tracing}. In \bibinfo{booktitle}{\emph{Proceedings of the 27th ACM SIGKDD conference on knowledge discovery \& data mining}}. \bibinfo{pages}{1452--1460}.
\newblock


\bibitem[Shen et~al\mbox{.}(2024)]%
        {shen2024survey}
\bibfield{author}{\bibinfo{person}{Shuanghong Shen}, \bibinfo{person}{Qi Liu}, \bibinfo{person}{Zhenya Huang}, \bibinfo{person}{Yonghe Zheng}, \bibinfo{person}{Minghao Yin}, \bibinfo{person}{Minjuan Wang}, {and} \bibinfo{person}{Enhong Chen}.} \bibinfo{year}{2024}\natexlab{}.
\newblock \showarticletitle{A survey of knowledge tracing: Models, variants, and applications}.
\newblock \bibinfo{journal}{\emph{IEEE Transactions on Learning Technologies}} (\bibinfo{year}{2024}).
\newblock


\bibitem[Song et~al\mbox{.}(2020)]%
        {Mpnet}
\bibfield{author}{\bibinfo{person}{Kaitao Song}, \bibinfo{person}{Xu Tan}, \bibinfo{person}{Tao Qin}, \bibinfo{person}{Jianfeng Lu}, {and} \bibinfo{person}{Tie-Yan Liu}.} \bibinfo{year}{2020}\natexlab{}.
\newblock \showarticletitle{Mpnet: Masked and permuted pre-training for language understanding}.
\newblock \bibinfo{journal}{\emph{Advances in neural information processing systems}}  \bibinfo{volume}{33} (\bibinfo{year}{2020}), \bibinfo{pages}{16857--16867}.
\newblock


\bibitem[Song et~al\mbox{.}(2022)]%
        {Bi-CLKT}
\bibfield{author}{\bibinfo{person}{Xiangyu Song}, \bibinfo{person}{Jianxin Li}, \bibinfo{person}{Qi Lei}, \bibinfo{person}{Wei Zhao}, \bibinfo{person}{Yunliang Chen}, {and} \bibinfo{person}{Ajmal Mian}.} \bibinfo{year}{2022}\natexlab{}.
\newblock \showarticletitle{Bi-CLKT: Bi-graph contrastive learning based knowledge tracing}.
\newblock \bibinfo{journal}{\emph{Knowledge-Based Systems}}  \bibinfo{volume}{241} (\bibinfo{year}{2022}), \bibinfo{pages}{108274}.
\newblock


\bibitem[Su et~al\mbox{.}(2018)]%
        {EERNN}
\bibfield{author}{\bibinfo{person}{Yu Su}, \bibinfo{person}{Qingwen Liu}, \bibinfo{person}{Qi Liu}, \bibinfo{person}{Zhenya Huang}, \bibinfo{person}{Yu Yin}, \bibinfo{person}{Enhong Chen}, \bibinfo{person}{Chris Ding}, \bibinfo{person}{Si Wei}, {and} \bibinfo{person}{Guoping Hu}.} \bibinfo{year}{2018}\natexlab{}.
\newblock \showarticletitle{Exercise-enhanced sequential modeling for student performance prediction}. In \bibinfo{booktitle}{\emph{Proceedings of the AAAI conference on artificial intelligence}}, Vol.~\bibinfo{volume}{32}.
\newblock


\bibitem[Sun et~al\mbox{.}(2024)]%
        {ProKT}
\bibfield{author}{\bibinfo{person}{Jianwen Sun}, \bibinfo{person}{Mengqi Wei}, \bibinfo{person}{Jintian Feng}, \bibinfo{person}{Fenghua Yu}, \bibinfo{person}{Qing Li}, {and} \bibinfo{person}{Rui Zou}.} \bibinfo{year}{2024}\natexlab{}.
\newblock \showarticletitle{Progressive knowledge tracing: Modeling learning process from abstract to concrete}.
\newblock \bibinfo{journal}{\emph{Expert Systems with Applications}}  \bibinfo{volume}{238} (\bibinfo{year}{2024}), \bibinfo{pages}{122280}.
\newblock


\bibitem[Tan et~al\mbox{.}(2022)]%
        {BiDKT}
\bibfield{author}{\bibinfo{person}{Weicong Tan}, \bibinfo{person}{Yuan Jin}, \bibinfo{person}{Ming Liu}, {and} \bibinfo{person}{He Zhang}.} \bibinfo{year}{2022}\natexlab{}.
\newblock \showarticletitle{BiDKT: Deep Knowledge Tracing with BERT}. In \bibinfo{booktitle}{\emph{Ad Hoc Networks and Tools for IT}}, \bibfield{editor}{\bibinfo{person}{Wei Bao}, \bibinfo{person}{Xingliang Yuan}, \bibinfo{person}{Longxiang Gao}, \bibinfo{person}{Tom~H. Luan}, {and} \bibinfo{person}{David Bong~Jun Choi}} (Eds.). \bibinfo{publisher}{Springer International Publishing}, \bibinfo{address}{Cham}, \bibinfo{pages}{260--278}.
\newblock
\showISBNx{978-3-030-98005-4}


\bibitem[Tiana et~al\mbox{.}(2021)]%
        {BEKT}
\bibfield{author}{\bibinfo{person}{Zejie Tiana}, \bibinfo{person}{Guangcong Zhengc}, \bibinfo{person}{Brendan Flanaganb}, \bibinfo{person}{Jiazhi Mic}, {and} \bibinfo{person}{Hiroaki Ogatab}.} \bibinfo{year}{2021}\natexlab{}.
\newblock \showarticletitle{BEKT: deep knowledge tracing with bidirectional encoder representations from transformers}. In \bibinfo{booktitle}{\emph{Proceedings of the 29th International Conference on Computers in Education}}, Vol.~\bibinfo{volume}{2}. \bibinfo{pages}{6--2}.
\newblock


\bibitem[Tong et~al\mbox{.}(2020)]%
        {tong2020exercise}
\bibfield{author}{\bibinfo{person}{Hanshuang Tong}, \bibinfo{person}{Yun Zhou}, {and} \bibinfo{person}{Zhen Wang}.} \bibinfo{year}{2020}\natexlab{}.
\newblock \showarticletitle{Exercise hierarchical feature enhanced knowledge tracing}. In \bibinfo{booktitle}{\emph{Artificial Intelligence in Education: 21st International Conference, AIED 2020, Ifrane, Morocco, July 6--10, 2020, Proceedings, Part II 21}}. Springer, \bibinfo{pages}{324--328}.
\newblock


\bibitem[Touvron et~al\mbox{.}(2023)]%
        {llama2}
\bibfield{author}{\bibinfo{person}{Hugo Touvron}, \bibinfo{person}{Louis Martin}, \bibinfo{person}{Kevin~R. Stone}, \bibinfo{person}{Peter Albert}, \bibinfo{person}{Amjad Almahairi}, \bibinfo{person}{Yasmine Babaei}, \bibinfo{person}{Nikolay Bashlykov}, \bibinfo{person}{Soumya Batra}, \bibinfo{person}{Prajjwal Bhargava}, \bibinfo{person}{Shruti Bhosale}, \bibinfo{person}{Daniel~M. Bikel}, \bibinfo{person}{Lukas Blecher}, \bibinfo{person}{Cristian~Cant{\'o}n Ferrer}, \bibinfo{person}{Moya Chen}, \bibinfo{person}{Guillem Cucurull}, \bibinfo{person}{David Esiobu}, \bibinfo{person}{Jude Fernandes}, \bibinfo{person}{Jeremy Fu}, \bibinfo{person}{Wenyin Fu}, \bibinfo{person}{Brian Fuller}, \bibinfo{person}{Cynthia Gao}, \bibinfo{person}{Vedanuj Goswami}, \bibinfo{person}{Naman Goyal}, \bibinfo{person}{Anthony~S. Hartshorn}, \bibinfo{person}{Saghar Hosseini}, \bibinfo{person}{Rui Hou}, \bibinfo{person}{Hakan Inan}, \bibinfo{person}{Marcin Kardas}, \bibinfo{person}{Viktor Kerkez}, \bibinfo{person}{Madian Khabsa},
  \bibinfo{person}{Isabel~M. Kloumann}, \bibinfo{person}{A.~V. Korenev}, \bibinfo{person}{Punit~Singh Koura}, \bibinfo{person}{Marie-Anne Lachaux}, \bibinfo{person}{Thibaut Lavril}, \bibinfo{person}{Jenya Lee}, \bibinfo{person}{Diana Liskovich}, \bibinfo{person}{Yinghai Lu}, \bibinfo{person}{Yuning Mao}, \bibinfo{person}{Xavier Martinet}, \bibinfo{person}{Todor Mihaylov}, \bibinfo{person}{Pushkar Mishra}, \bibinfo{person}{Igor Molybog}, \bibinfo{person}{Yixin Nie}, \bibinfo{person}{Andrew Poulton}, \bibinfo{person}{Jeremy Reizenstein}, \bibinfo{person}{Rashi Rungta}, \bibinfo{person}{Kalyan Saladi}, \bibinfo{person}{Alan Schelten}, \bibinfo{person}{Ruan Silva}, \bibinfo{person}{Eric~Michael Smith}, \bibinfo{person}{R. Subramanian}, \bibinfo{person}{Xia Tan}, \bibinfo{person}{Binh Tang}, \bibinfo{person}{Ross Taylor}, \bibinfo{person}{Adina Williams}, \bibinfo{person}{Jian~Xiang Kuan}, \bibinfo{person}{Puxin Xu}, \bibinfo{person}{Zhengxu Yan}, \bibinfo{person}{Iliyan Zarov}, \bibinfo{person}{Yuchen Zhang},
  \bibinfo{person}{Angela Fan}, \bibinfo{person}{Melanie Kambadur}, \bibinfo{person}{Sharan Narang}, \bibinfo{person}{Aurelien Rodriguez}, \bibinfo{person}{Robert Stojnic}, \bibinfo{person}{Sergey Edunov}, {and} \bibinfo{person}{Thomas Scialom}.} \bibinfo{year}{2023}\natexlab{}.
\newblock \showarticletitle{Llama 2: Open Foundation and Fine-Tuned Chat Models}.
\newblock \bibinfo{journal}{\emph{ArXiv}}  \bibinfo{volume}{abs/2307.09288} (\bibinfo{year}{2023}).
\newblock
\urldef\tempurl%
\url{https://api.semanticscholar.org/CorpusID:259950998}
\showURL{%
\tempurl}


\bibitem[Vaswani(2017)]%
        {vaswani2017attention}
\bibfield{author}{\bibinfo{person}{Ashish Vaswani}.} \bibinfo{year}{2017}\natexlab{}.
\newblock \showarticletitle{Attention is all you need}.
\newblock \bibinfo{journal}{\emph{arXiv preprint arXiv:1706.03762}} (\bibinfo{year}{2017}).
\newblock


\bibitem[Wang et~al\mbox{.}(2021)]%
        {HawkesKT}
\bibfield{author}{\bibinfo{person}{Chenyang Wang}, \bibinfo{person}{Weizhi Ma}, \bibinfo{person}{Min Zhang}, \bibinfo{person}{Chuancheng Lv}, \bibinfo{person}{Fengyuan Wan}, \bibinfo{person}{Huijie Lin}, \bibinfo{person}{Taoran Tang}, \bibinfo{person}{Yiqun Liu}, {and} \bibinfo{person}{Shaoping Ma}.} \bibinfo{year}{2021}\natexlab{}.
\newblock \showarticletitle{Temporal cross-effects in knowledge tracing}. In \bibinfo{booktitle}{\emph{Proceedings of the 14th ACM International Conference on Web Search and Data Mining}}. \bibinfo{pages}{517--525}.
\newblock


\bibitem[Wang et~al\mbox{.}(2020)]%
        {Nips2020}
\bibfield{author}{\bibinfo{person}{Zichao Wang}, \bibinfo{person}{Angus Lamb}, \bibinfo{person}{Evgeny Saveliev}, \bibinfo{person}{Pashmina Cameron}, \bibinfo{person}{Yordan Zaykov}, \bibinfo{person}{Jos{\'e}~Miguel Hern{\'a}ndez-Lobato}, \bibinfo{person}{Richard~E Turner}, \bibinfo{person}{Richard~G Baraniuk}, \bibinfo{person}{Craig Barton}, \bibinfo{person}{Simon~Peyton Jones}, \bibinfo{person}{Simon Woodhead}, {and} \bibinfo{person}{Cheng Zhang}.} \bibinfo{year}{2020}\natexlab{}.
\newblock \showarticletitle{Diagnostic questions: The neurips 2020 education challenge}.
\newblock \bibinfo{journal}{\emph{arXiv preprint arXiv:2007.12061}} (\bibinfo{year}{2020}).
\newblock


\bibitem[Xu et~al\mbox{.}(2023)]%
        {LBKT}
\bibfield{author}{\bibinfo{person}{Bihan Xu}, \bibinfo{person}{Zhenya Huang}, \bibinfo{person}{Jiayu Liu}, \bibinfo{person}{Shuanghong Shen}, \bibinfo{person}{Qi Liu}, \bibinfo{person}{Enhong Chen}, \bibinfo{person}{Jinze Wu}, {and} \bibinfo{person}{Shijin Wang}.} \bibinfo{year}{2023}\natexlab{}.
\newblock \showarticletitle{Learning behavior-oriented knowledge tracing}. In \bibinfo{booktitle}{\emph{Proceedings of the 29th ACM SIGKDD conference on knowledge discovery and data mining}}. \bibinfo{pages}{2789--2800}.
\newblock


\bibitem[Zanellati et~al\mbox{.}(2024)]%
        {zanellati2024hybrid}
\bibfield{author}{\bibinfo{person}{Andrea Zanellati}, \bibinfo{person}{Daniele Di~Mitri}, \bibinfo{person}{Maurizio Gabbrielli}, {and} \bibinfo{person}{Olivia Levrini}.} \bibinfo{year}{2024}\natexlab{}.
\newblock \showarticletitle{Hybrid models for knowledge tracing: A systematic literature review}.
\newblock \bibinfo{journal}{\emph{IEEE Transactions on Learning Technologies}} (\bibinfo{year}{2024}).
\newblock


\bibitem[Zhang et~al\mbox{.}(2017)]%
        {DKVMN}
\bibfield{author}{\bibinfo{person}{Jiani Zhang}, \bibinfo{person}{Xingjian Shi}, \bibinfo{person}{Irwin King}, {and} \bibinfo{person}{Dit-Yan Yeung}.} \bibinfo{year}{2017}\natexlab{}.
\newblock \showarticletitle{Dynamic Key-Value Memory Networks for Knowledge Tracing}. In \bibinfo{booktitle}{\emph{Proceedings of the 26th International Conference on World Wide Web}}.
\newblock
\urldef\tempurl%
\url{https://doi.org/10.1145/3038912.3052580}
\showDOI{\tempurl}


\end{thebibliography}

\appendix

\begin{table*}[hbt!]
\centering
\small
\begin{tabular}{| >{\centering\arraybackslash}m{0.1\textwidth} | >{\raggedright\arraybackslash}m{0.8\textwidth} |}
\hline
\rowcolor{lightgray} 
\textbf{Type} & \multicolumn{1}{c|}{\textbf{Template}} \\
\hline
1 & The student has previously, in chronological order, answered question with ID=74 [WrapQEmb]
involving concept ID=6 [WrapCEmb] correctly, ...,
question with ID=42 [WrapQEmb] involving concept ID=5 [WrapCEmb] incorrectly.
Please predict whether the student will answer the next question with ID=44 [NextWrapQEmb]
involving concept ID=5 [NextWrapCEmb] correctly. 
Response with ‘Yes’ or ‘No’. \\
\hline
2 & The student has previously, in chronological order, answered question with ID=3117 [QidEmb]
correctly, question with ID=2964 [QidEmb] correctly, 
question with ID=5627 [QidEmb] incorrectly, ...,
question with ID=5532 [QidEmb] correctly.
Please predict whether the student will answer the next question with ID=5707 [NextQidEmb] correctly.
Response with ‘Yes’ or ‘No’. \\
\hline
3 & The student has previously, in chronological order, answered question involving concept ID=15 [CidEmb]
correctly, ...,
question involving concept ID=30 [NextCidEmb] correctly.
Please predict whether the student will answer the next question involving concept ID=30 correctly.
Response with ‘Yes’ or ‘No’. \\
\hline
\end{tabular}
\caption{The prompt templates for LLM-KT(Ours)}
\label{table:llm-kt-templates}
\end{table*}

\begin{table*}[hbt!]
\centering
\small
\begin{tabular}{| >{\centering\arraybackslash}m{0.1\textwidth} | >{\raggedright\arraybackslash}m{0.8\textwidth} |}
\hline
\rowcolor{lightgray} 
\textbf{Type} & \multicolumn{1}{c|}{\textbf{Template}} \\
\hline
1 & The student has previously, in chronological order, answered question with ID=74
involving concept ID=6 correctly, question with ID=80 involving concept ID=6 correctly,
...,
question with ID=42 involving concept ID=5 incorrectly.
Please predict whether the student will answer the next question with ID=44
involving concept ID=5 correctly.
Response with ‘Yes’ or ‘No’. \\
\hline
2 & The student has previously, in chronological order, answered question with ID=3117
correctly, question with ID=2964 correctly,
...,
question with ID=5532 correctly.
Please predict whether the student will answer the next question with ID=5707 correctly.
Response with ‘Yes’ or ‘No’. \\
\hline
3 & The student has previously, in chronological order, answered question involving concept ID=15
correctly, question involving concept ID=15 correctly,
...,
question involving concept ID=30 correctly.
Please predict whether the student will answer the next question involving concept ID=30 correctly.
Response with ‘Yes’ or ‘No’. \\
\hline
\end{tabular}
\caption{The prompt templates for LLM-FT$_\mathrm{ID}$}
\label{table:llm-kt-id-templates}
\end{table*}

\begin{table*}[hbt!]
\centering
\small
\begin{tabular}{| >{\centering\arraybackslash}m{0.1\textwidth} | >{\raggedright\arraybackslash}m{0.8\textwidth} |}
\hline
\rowcolor{lightgray} 
\textbf{Type} & \multicolumn{1}{c|}{\textbf{Template}} \\
\hline
1 & The student has previously, in chronological order, answered question with ID=[qid$_{74}$] involving concept ID=[cid$_{6}$] correctly, question with ID=[qid$_{80}$] involving concept ID=[cid$_{6}$] correctly, ..., question with ID=[qid$_{42}$] involving concept ID=[cid$_{5}$] incorrectly. Please predict whether the student will answer the next question with ID=[qid$_{44}$] involving concept ID=[cid$_{5}$] correctly. Response with `Yes' or `No'. \\
\hline
2 & The student has previously, in chronological order, answered question with ID=[qid$_{3117}$] correctly, question with ID=[qid$_{2964}$] correctly, question with ID=[qid$_{5627}$] incorrectly, ..., question with ID=[qid$_{5532}$] correctly. Please predict whether the student will answer the next question with ID=[qid$_{5707}$] correctly. Response with `Yes' or `No'. \\
\hline
3 & The student has previously, in chronological order, answered question involving concept ID=[cid$_{15}$] correctly, question involving concept ID=[cid$_{15}$] correctly, question involving concept ID=[cid$_{30}$] incorrectly, ..., question involving concept ID=[cid$_{30}$] correctly. Please predict whether the student will answer the next question involving concept ID=[cid$_{30}$] correctly. Response with `Yes' or `No'. \\
\hline
\end{tabular}
\caption{The prompt templates for LLM-FT$_\mathrm{TokenID}$}
\label{table:llm-kt-tokenid-templates}
\end{table*}

\begin{table*}[hbt!]
\centering
\small
\begin{tabular}{| >{\centering\arraybackslash}m{0.1\textwidth} | >{\raggedright\arraybackslash}m{0.8\textwidth} |}
\hline
\rowcolor{lightgray} 
\textbf{Type} & \multicolumn{1}{c|}{\textbf{Template}} \\
\hline
4 & In this task, we aim to determine whether the student can answer the question correctly based on the student's history record of academic exercises. \newline
    The student's history record of academic exercises is given as follows: \newline
    1) How would this calculation be written? Pic$_{290-0}$ \newline
    A:8+(2÷5)=2 B:(8+2)÷5=2 C:8+2÷5=2 D:(8+2÷5)=2 \newline
    Related knowledge concepts: Basic Arithmetic \newline
    The student answered this question correctly \newline
    2) Which symbol belongs in the box? Pic$_{749-0}$ \newline
    A:$>$  B:$<$  C:$=$ D:$\ge$ \newline
    Related knowledge concepts: Basic Arithmetic \newline
    The student answered this question correctly \newline
    3) What is the output of this Function Machine? Pic$_{836-0}$ \newline
    A:10p B:7p  C:5(p+2)  D:5p+2 \newline
    Related knowledge concepts: Writing Expressions \newline
    The student answered this question incorrectly \newline
    The target question is given as follows: \newline
    Tom and Katie are arguing about the result of this Function Machine: Pic$_{856-0}$. Tom says the output is: 3n-12. Katie says the output is:3(n-4). Who is correct? \newline
    A:Only Tom B:Only Katie C:Both Tom and Katie  D:Neither is correct \newline
    Related knowledge concepts: Writing Expressions \newline
    Please predict whether the student would answer the target question correctly. Response with `Yes' or `No'. \\
\hline
5 & The student has previously, in chronological order, answered question involving concept ``Basic Arithmetic" correctly, question involving concept ``Basic Arithmetic" correctly, ..., question involving concept ``Basic Arithmetic" incorrectly, question involving concept ``Basic Arithmetic" correctly, ..., question involving concept ``Ordering Negative Numbers" incorrectly, question involving concept ``Ordering Negative Numbers" correctly. Please predict whether the student will answer the next question involving concept ``Ordering Negative Numbers" correctly. Response with `Yes' or `No'. \\
\hline
\end{tabular}
\caption{The prompt templates for LLM-FT$_\mathrm{Text}$}
\label{table:llm-kt-text-templates}
\end{table*}

\section{Prompt Templates}

In this section, we provide detailed descriptions of the prompt templates used for different datasets in our study. These templates are designed to handle various types of data and adapt to the specific requirements of each dataset. 

We introduce five distinct prompt templates:
\begin{itemize}
    \item \textbf{Type 1} (Combined Question and Concept ID Template): Used for datasets with both QIDs and CIDs, applicable to Assist2009 and Nips2020.
    \item \textbf{Type 2} (Question ID-Only Template): Used exclusively for datasets with only QIDs, such as Junyi.
    \item \textbf{Type 3} (Concept ID-Only Template): Used exclusively for datasets with only CIDs, like Assist2015.
    \item \textbf{Type 4} (Contextual Question Template): Used for datasets with text associated with questions, applicable only to Nips2020.
    \item \textbf{Type 5} (Contextual Concept Template): Used for datasets with concept text, like Assist2009.
\end{itemize}

\section{Terminology Explanation}

\begin{itemize}
    \item \textbf{QID} (Question ID): The unique identifier for each question, used to track and model the sequence of a student’s answers. 
    \item \textbf{CID} (Concept ID): The unique identifier for the knowledge concept tied to each question.
    \item \textbf{WrapQEmb} (Wrapped Question Embedding): The embedding formed by combining the QID and the question’s text, leveraging both identity and semantic content.
    \item \textbf{WrapCEmb} (Wrapped Concept Embedding): Similar to ‘WrapQEmb’, but combines the CID with the concept’s text.
    \item \textbf{QidEmb} (Question ID Embedding): An embedding of the QID, used without the question’s text in templates focused on identity.
    \item \textbf{CidEmb} (Concept ID Embedding): An embedding of the CID, used without the concept’s text in simpler templates.
    \item \textbf{NextWrapQEmb} (Next Wrapped Question Embedding): The fused embedding for the next QID, combining its ID and text similar to ‘WrapQEmb’.
    \item \textbf{NextQidEmb} (Next Question ID Embedding): The next QID’s embedding, used without the question’s text.
    \item \textbf{NextWrapCEmb} (Next Wrapped Concept Embedding): The fused embedding for the next CID, combining its ID and text.
    \item \textbf{NextCidEmb} (Next Concept ID Embedding): The next CID’s embedding, used without the concept’s text.
\end{itemize}

\end{document}